\documentclass[runningheads]{llncs}

\makeatletter
\ifdefined\pdfcompresslevel\else\let\pdfcompresslevel\@tempcnta\fi
\ifdefined\pdfoptionpdfminorversion\else\let\pdfoptionpdfminorversion\@tempcnta\fi
\ifdefined\pdfglyphtounicode\else\def\pdfglyphtounicode#1#2{}\fi
\ifdefined\pdfgentounicode\else\let\pdfgentounicode\@tempcnta\fi
\makeatother

\PassOptionsToPackage{table}{xcolor}
\usepackage{eccv}
\usepackage{eccvabbrv}

\usepackage{graphicx}
\usepackage{booktabs}
\usepackage{url}
\usepackage{float}
\usepackage{makecell}
\usepackage{multirow}
\usepackage{tabularx}
\usepackage{kotex}
\usepackage{threeparttable}
\usepackage{wrapfig}
\usepackage{array}
\usepackage{subcaption}
\usepackage{amsmath, amssymb}
\usepackage{tikz}
\usetikzlibrary{tikzmark,calc}
\usepackage[accsupp]{axessibility}

\usepackage[colorlinks,urlcolor=eccvblue,linkcolor=eccvblue,citecolor=eccvblue]{hyperref}
\usepackage{orcidlink}

\begin{document}

\title{OSVE: One Step Video Editing \\ with One Step Diffusion Models}

\author{Habin Lim\orcidlink{0009-0001-5543-424X} \and
Gyeong-Moon Park\orcidlink{0000-0003-4011-9981}\thanks{Corresponding author.}}
\authorrunning{H.~Lim and G.-M.~Park}
\institute{Korea University, Seoul, Republic of Korea\\
\email{\{ha001211,gm-park\}@korea.ac.kr}}

\newcommand{\Ours}{OSVE}

\maketitle
\begin{abstract}
Text-guided video editing with diffusion models is impractically slow, hindered by costly multi-step sampling and inversion. We present OSVE, the first framework to successfully adapt one-step Text-to-Image (T2I) models for high-quality video editing, addressing the core challenges of inversion, editability, and temporal consistency. To bypass slow iterative inversion, we train a learnable encoder that predicts the initial noise for each frame in a single forward pass. This encoder is trained with a novel Structure-Aware Editing (SAE) loss on a curated dataset of structurally-aligned image pairs, teaching it to preserve the source video's geometry during edits. For temporal coherence, we introduce Unified-Frame Editing (UFE), a technique that concatenates frame latents to facilitate cross-frame attention in a single generation step. Furthermore, for long videos, a sliding-window strategy with an anchor frame maintains global consistency. Our extensive experiments demonstrate that OSVE achieves editing quality comparable or superior to state-of-the-art multi-step methods, while operating approximately 155--171 times faster. This breakthrough paves the way for practical, real-time video editing applications. Code is available at \url{https://github.com/KU-VGI/OSVE}.
\end{abstract}
\section{Introduction}
\label{sec:intro}

Text-guided video editing, \ie, modifying a source video's visual content according to a natural-language instruction, is increasingly vital for content creation, filmmaking, and interactive media.
Recent generative diffusion models, spanning Text-to-Image (T2I)~\cite{pmlr-v37-sohl-dickstein15,ho2020denoising,song2020score,rombach2022high} and Text-to-Video (T2V)~\cite{wan2025wanopenadvancedlargescale,hacohen2024ltxvideorealtimevideolatent,bartal2024lumiere,openai2024sora,kong2024hunyuanvideo} architectures, have enabled the generation of photorealistic, temporally coherent visual content from text alone.
Building upon these powerful backbones, a growing body of work performs training-free, zero-shot video editing by intervening in the diffusion process~\cite{wu2023tuneavideooneshottuningimage,geyer2023tokenflow,cong2024flattenopticalflowguidedattention,liu2024video,kara2024rave,wang2024cove,wang2025videodirector,jiang2025vace}.
These methods typically follow a two-stage pipeline: (i)~\textit{inverting} the source video via multi-step inversion~\cite{song2020ddim,mokady2023null} to recover its latent trajectory, and (ii)~\textit{generating} the edited frames by re-running the diffusion process under a modified text prompt, often with cross-frame attention control~\cite{tumanyan2023plug,hertz2023prompttoprompt} to preserve spatial structure and temporal layout.

Despite the editing quality these methods achieve, they inherit a critical bottleneck: \textbf{extreme computational cost}.
Because every frame must be processed over tens to hundreds of denoising steps, wall-clock time scales with both the number of frames and the number of steps.
For instance, RAVE~\cite{kara2024rave}, reported as one of the fastest T2I-based editing methods, still requires approximately 24 hours to edit a 5-minute, 30\,FPS video\footnote{Measured on a single NVIDIA RTX 6000 Ada GPU at $512\times512$ resolution.}.
T2V backbones, which incorporate heavyweight temporal attention modules, demand even greater resources.
This extreme latency renders current approaches impractical for real-time or streaming-style applications~\cite{xing2024survey}.

\begin{figure*}[t]
    \centering
    \includegraphics[width=0.95\linewidth]{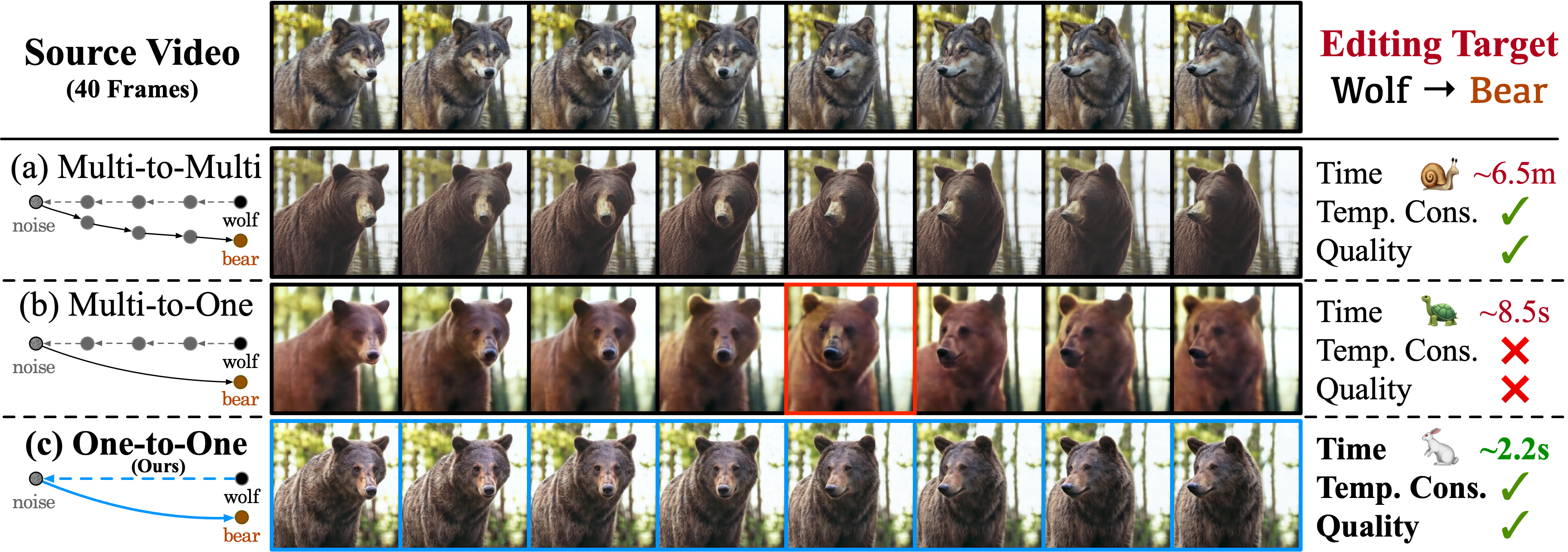}
    \caption{\textbf{Three frameworks for text-guided video editing.} \textit{(a) Multi-to-Multi:} The standard approach yields high-quality edits but is impractically slow. \textit{(b) Multi-to-One:} Na\"ively adopting a one-step diffusion model produces degraded, temporally inconsistent outputs. \textit{(c) One-to-One (Ours):} Our proposed framework uses a learnable one-step encoder to enable fast, temporally consistent, and high-quality video editing, resolving the speed--quality trade-off.}
    \label{fig:motivation}
    \vspace{-2mm}
\end{figure*}

To overcome this bottleneck, we shift toward a \textit{one-step} generation paradigm.
One-step diffusion models~\cite{yin2024improved,dao2024swiftbrush,nguyen2024swiftbrush,yin2024one,liu2023instaflow}, distilled from their multi-step teachers, have achieved competitive or even superior image quality, yet their potential for video editing remains unexplored.
By adapting this ultra-fast regime to video, we build a One-to-One editing framework (one-step inversion followed by one-step generation) that replaces the costly iterative denoising loop with a single forward pass.
We instantiate and validate our framework on a publicly released one-step T2I backbone (DMD2~\cite{yin2024improved}) to demonstrate its effectiveness with today's most accessible one-step generators.

However, directly applying one-step diffusion models to video editing presents three critical challenges (see Figure~\ref{fig:motivation}(b)):

\noindent\textbf{(i) Inversion--generation mismatch.}
Conventional multi-step inversion~\cite{song2020ddim} is both \textit{time-consuming} and fundamentally ill-suited for one-step generators.
Since the inversion follows a multi-step trajectory while the generator performs a direct mapping, the mismatch between forward and reverse paths causes severe information loss, producing blurred and degraded results.

\noindent\textbf{(ii) Structural collapse under single-step control.}
Preserving the source video's spatial structure (\eg, a wolf's rotating head) while altering its appearance (\eg, to a bear) is critical for faithful editing.
We empirically find that existing structure-preserving methods such as Prompt-to-Prompt~\cite{hertz2023prompttoprompt} and ControlNet~\cite{zhang2023adding} fail when constrained to a single step.
Unlike multi-step processes where control signals are distributed across steps and gradually absorbed, one-step models concentrate the same signal into a single pass with no room for correction, causing it to dominate rather than guide the output, a phenomenon we term ``over-steering''.

\noindent\textbf{(iii) Temporal inconsistency.}
One-step T2I diffusion models do not inherently encode temporal information, resulting in flickering or discontinuous frames when applied to video.
Existing temporal consistency techniques, including attention fine-tuning~\cite{wu2023tuneavideooneshottuningimage}, cross-frame attention control~\cite{qu2024tokenflowunifiedimagetokenizer,cong2024flattenopticalflowguidedattention,liu2024video,wang2024cove}, and noise shuffling across sampling steps~\cite{kara2024rave}, all rely on the multi-step structure and are therefore incompatible with the one-step regime.

To address these challenges, we propose \textbf{OSVE} (\textbf{O}ne-\textbf{S}tep \textbf{V}ideo \textbf{E}diting), a One-to-One video editing framework explicitly designed for one-step generation.
For \textbf{inversion--generation mismatch}, the core of OSVE is a \textit{learnable video encoder}, initialized from a pre-trained U-Net to inherit a strong prior over the generator's latent space, that replaces slow multi-step inversion with a single forward pass.
For \textbf{structural collapse}, we introduce a \textbf{Structure-Aware Editing (SAE)} loss that trains the encoder to produce latents inherently preserving structural integrity under modified text prompts, eliminating the need for external guidance that fails in one step. By training on diverse editing pairs, the encoder learns structure-aware latent representations that maintain the source's layout while permitting appearance changes (Figure~\ref{fig:motivation}(c)).
For \textbf{temporal inconsistency}, we propose \textbf{Unified-Frame Editing (UFE)}, which concatenates the inverted latents of consecutive frames into a single unified tensor, enabling the model's self-attention to align features across frames in one pass. Combined with a sliding-window strategy for local coherence and anchor-frame selection for long-range stability, UFE ensures both short-term and long-term temporal consistency.

\vspace{1mm}
\noindent In summary, our main contributions are:
\begin{itemize}
    \item We present \textbf{OSVE}, the first One-to-One video editing framework for the one-step diffusion regime, replacing multi-step inversion with a learnable single-pass encoder.
    \item We propose a \textbf{Structure-Aware Editing (SAE) loss} that directly injects structural preservation into the inversion stage, mitigating the over-steering problem inherent in single-step control.
    \item We introduce \textbf{Unified-Frame Editing (UFE)}, which enforces temporal consistency via unified latent processing with a sliding window and anchor frame, ensuring both short- and long-range coherence.
    \item Extensive experiments show that OSVE achieves \textbf{sub-second latency}, $\sim 155\text{--}171\times$ faster than the previous fastest method, while delivering comparable or superior visual quality, enabling practical real-time video editing.
\end{itemize}

\section{Related Work}
\subsubsection{Text-Guided Video Editing with Diffusion Models.}
Building on advances in image-level generative editing, including GAN inversion~\cite{tov2021designing, roich2022pivotal, moon2022interestyle, kim2025wine} and diffusion-based manipulation~\cite{hertz2023prompttoprompt, mokady2023null, brooks2022instructpix2pix, lim2025conceptsplit}, numerous zero-shot techniques have been proposed to extend such capabilities to video while preserving temporal consistency. These include fusing attention maps~\cite{qi2023fatezero}, propagating token features~\cite{geyer2023tokenflow, wang2024cove}, using optical flow guidance~\cite{cong2024flattenopticalflowguidedattention}, and shuffling noise~\cite{kara2024rave}. While these approaches have significantly improved video coherence, they all operate within a multi-step sampling framework. This iterative nature, where operations are repeated for each frame across numerous steps, results in prohibitively long runtimes. This computational cost remains a critical barrier, motivating our work on high-quality, single-step video editing.
\vspace{-5pt}
\subsubsection{One-Step Diffusion Models.}
To overcome the significant latency of iterative diffusion sampling, a prominent line of research focuses on distilling multi-step teacher models into highly efficient one- or few-step generators. This paradigm was advanced by methods like Progressive Distillation and Consistency Models, with latent variants (LCM/LCM-LoRA) enabling 2--8 step generation for Stable Diffusion~\cite{salimans2022progressive,song2023consistency,luo2023latent}. Subsequent works have further refined this approach. Rectified Flow and its successor, InstaFlow, straighten probability-flow trajectories to achieve SD-level quality in a single step via reflow-based distillation~\cite{liu2023flow, liu2023instaflow}. Other strategies, such as Adversarial Diffusion Distillation (\eg, SDXL-Turbo, SDXL-Lightning) and various distribution matching techniques (\eg, DMD, SwiftBrush), couple teacher supervision with GAN losses or novel distillation objectives to achieve high-quality synthesis in just 1--4 steps~\cite{sauer2024adversarial,lin2024sdxllightning,yin2024one,yin2024improved,nguyen2024swiftbrush,dao2024swiftbrush}.
These one-step models have revolutionized image generation speed, but this progress has been exclusively image-centric. This creates a critical gap and a major opportunity, as the domain of video editing desperately needs such acceleration. Our work is the first to bridge this gap, introducing the necessary mechanisms to successfully adapt these fast generators for video, thereby unlocking their potential for practical, high-speed applications.
\begin{figure*}[htbp]
    \centering
    \includegraphics[width=0.98\linewidth]{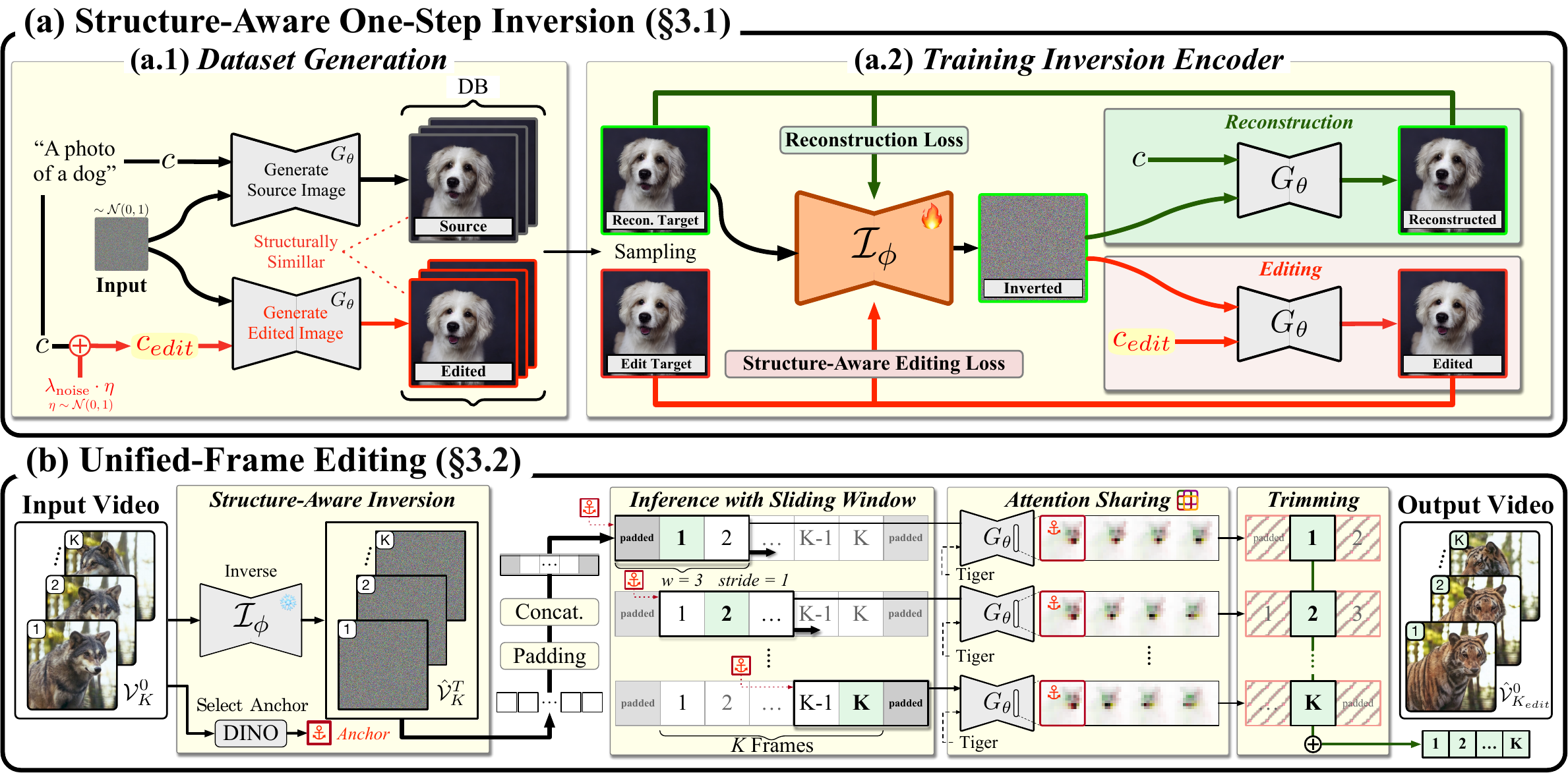}
    \caption{\textbf{Overview of our OSVE framework for one-step video editing.} \textit{(a) Training}: We train an inversion encoder using a novel dataset of structurally aligned image pairs generated via \textit{Prompt Perturbation}. The encoder learns to predict an initial noise that supports both faithful reconstruction and structure-aware editing. \textit{(b) Inference}: Our \textit{Unified-Frame Editing (UFE)} method inverts all video frames, concatenates their latents into a single map, and processes them in one pass. This enables cross-frame attention, ensuring temporal consistency. An anchor-based sliding window scales the approach to long videos while maintaining global coherence.}
    \label{fig:mainfigure}
    \vspace{-10pt}
\end{figure*}
\section{OSVE: One-Step Video Editing}

We present \Ours, a framework for one-step video editing built on two core ideas: (i) encoding source structure into the initial noise latent so that a single-pass generator can preserve it without iterative control, and (ii) enforcing temporal consistency by letting all frames share context through a unified attention mechanism.
Section~\ref{sec:3.1} develops the first idea through a structure-aware inversion encoder trained with a dedicated editing loss.
Section~\ref{sec:3.2} introduces Unified-Frame Editing (UFE) for the second, along with sliding-window and anchor-based schemes that scale it to long videos.
Figure~\ref{fig:mainfigure} illustrates an overview of the framework.

\subsection{Structure-Aware One-Step Inversion}
\label{sec:3.1}

Recent one-step diffusion models have dramatically accelerated image and video generation, yet their application to editing remains severely limited.
The core difficulty is \textbf{structure preservation}: an edit should maintain the spatial layout, object pose, and geometric relationships of the source while modifying only the intended semantics.
Structure preservation has been the central concern in diffusion-based editing from the outset, as methods such as Prompt-to-Prompt (P2P)~\cite{hertz2023prompttoprompt}, ControlNet~\cite{zhang2023adding}, and PnP~\cite{tumanyan2023plug} were all designed primarily to keep the source structure intact during semantic modification.
Without it, edits suffer from spatial misalignment and temporal inconsistency (Figure~\ref{fig:motivation}(b)), rendering them practically unusable regardless of semantic quality.
While multi-step diffusion models have largely addressed this through iterative inversion and control, one-step models lack such mechanisms, making structure preservation the central bottleneck that must be solved before one-step editing becomes practical.

To understand why these existing solutions do not transfer, consider how they work in the multi-step setting.
DDIM inversion~\cite{song2020ddim} traces the denoising trajectory backward to recover a latent that approximately reconstructs the source, and the editing techniques listed above then intervene at selected steps along this trajectory.
Crucially, these methods rely on the iterative nature of the process: each denoising step provides a corrective feedback loop that absorbs small perturbations introduced by the control signal, gradually steering the generation while keeping the structure intact.
One-step models, however, collapse the entire trajectory into a single forward pass, which creates two coupled difficulties.
First, there is no multi-step trajectory to reverse, so standard DDIM-style inversion is inapplicable.
Second, and more fundamentally, iterative control mechanisms lose the corrective loop they depend on.
In iterative generation, the control signal is distributed across many denoising steps, and each step applies only a small perturbation that subsequent steps can absorb and correct.
When the same control is concentrated into a single pass, its full influence is applied at once with no opportunity for subsequent correction, causing the control signal to dominate the generation rather than gently guide it.
The resulting ``over-steering'' artifacts are shown in Figure~\ref{fig:attentioninjection}(b, c, e, f).

This analysis points to a clear design principle: in a one-step setting, structure preservation cannot be imposed during generation but should be encoded before generation begins.
If the initial noise latent itself already carries the structural information of the source frame, the one-step generator can preserve that structure in its single forward pass without any auxiliary control mechanism.
We realize this principle by training an inversion encoder $\mathcal{I}_\phi(\cdot)$ that directly maps a source frame to an initial noise latent encoding the spatial structure needed for both faithful reconstruction and meaningful editing.

Concretely, given a VAE-encoded latent $z^0$ from a pre-trained autoencoder~\cite{kingma2014autoencoding,rombach2022high} and a prompt embedding $c = E(y)$ for a given text prompt $y$ (where $E$ is the frozen text encoder, \eg, CLIP~\cite{radford2021clip}), our encoder predicts an initial noise latent $\hat{z}^T = \mathcal{I}_\phi(z^0, c)$, where $T$ denotes the terminal diffusion timestep (\ie, 999).
Following previous inversion methods~\cite{song2020ddim,mokady2023null,hertz2023prompttoprompt,tumanyan2023plug,roich2022pivotal,tov2021designing}, our objective is to predict a noise latent $\hat{z}^T$ that simultaneously satisfies two criteria: \textbf{reconstruction} and \textbf{editability}.
To endow the encoder with structure-aware editability, we proceed in two stages:
\textbf{1) Dataset generation with prompt perturbation}: We curate a dataset of structurally similar image pairs that share pose and layout but differ visually (Figure~\ref{fig:mainfigure}(a.1)).
\textbf{2) Training with Structure-Aware Editing loss}: We train the encoder using a newly proposed structure-aware editing loss that simulates editing and constrains the output to align with the paired target (Figure~\ref{fig:mainfigure}(a.2)).

\vspace{-10pt}
\subsubsection{Dataset Generation with Prompt Perturbation.}
Training the inversion encoder to encode structure before generation requires data that teaches what ``structure preservation under semantic change'' looks like.
Specifically, we need pairs of images that share spatial layout but differ in appearance, so that the encoder learns to produce latents capturing the shared structure while remaining agnostic to the varying semantics.

\begin{figure*}[t]
  \centering
  \begin{tabular}{@{}c@{\hspace{1em}}c@{}}
    \begin{minipage}[c]{0.45\linewidth}
      \centering
      \small
      
      \captionof{table}{Quantitative comparison of reconstruction
      quality for different inversion strategies.}
      \vspace{10pt}
      \resizebox{\linewidth}{!}{%
      {\renewcommand{\arraystretch}{1.4}
        \begin{tabular}{lcccc}
          \toprule
           & {\scriptsize PSNR}$\uparrow$ & {\scriptsize LPIPS}$\downarrow$
           & {\scriptsize MSE}$\downarrow$ & {\scriptsize SSIM}$\uparrow$ \\
          \midrule
          DDIM~\cite{song2020ddim}      & 13.87 & 0.525 & 0.045 & 0.446 \\
          Off-the-shelf\textsuperscript{\ref{fn:offtheshelf}}
                                         & 17.54 & 0.400 & 0.032 & 0.543 \\
          \rowcolor{gray!15}
          \textbf{Ours} & \textbf{19.16} & \textbf{0.261}
                        & \textbf{0.014} & \textbf{0.624} \\
          \bottomrule
        \end{tabular}%
      }}
      \label{tab:off-the-shelf}
    \end{minipage}
    &
    \begin{minipage}[c]{0.53\linewidth}
      \centering
      \includegraphics[width=\linewidth]{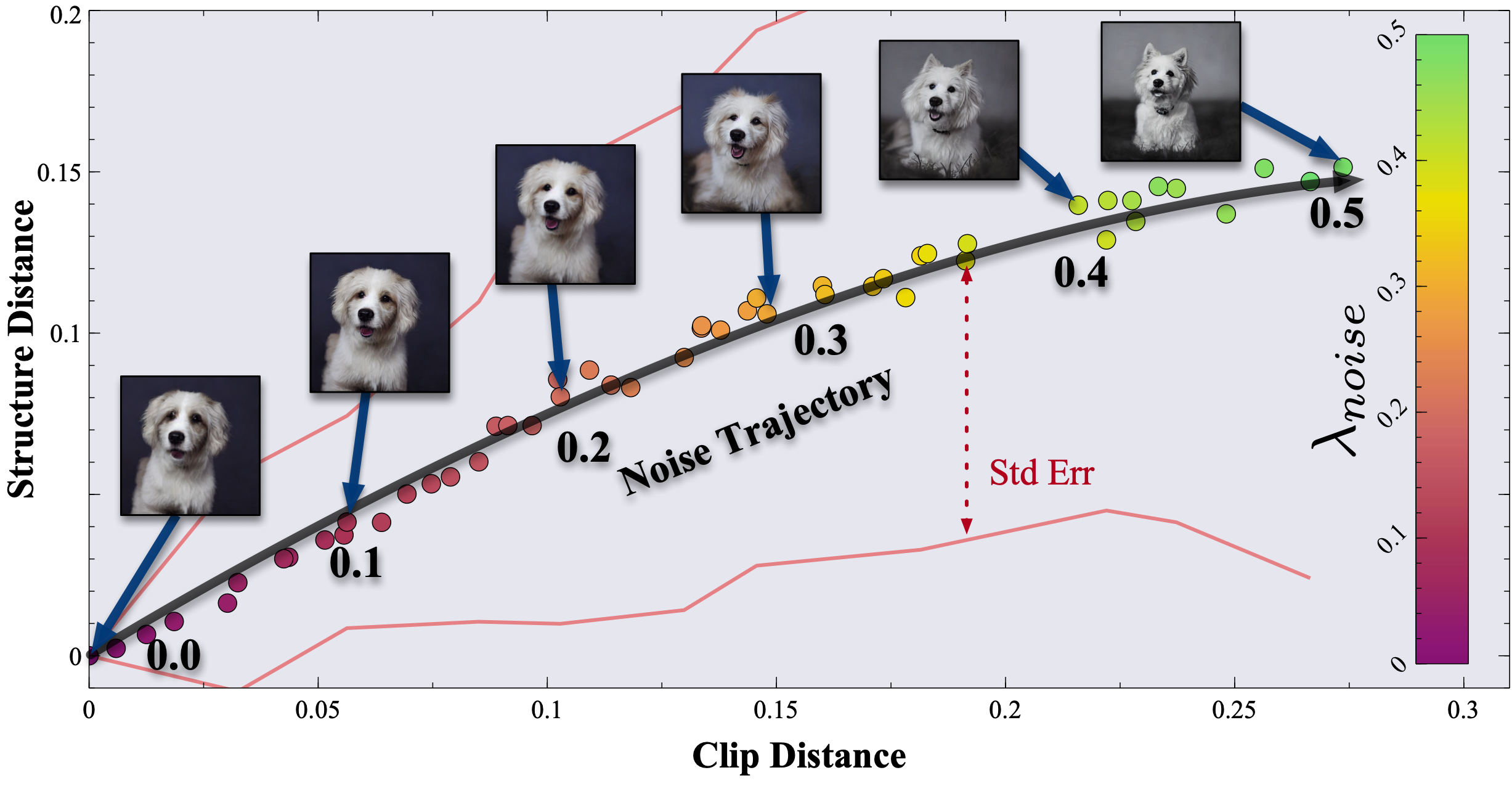}
      \vspace{-15pt}
      \captionof{figure}{Structural (DINO) and semantic (CLIP)
      distances as $\lambda_{\mathrm{noise}}$ varies.}
      \label{fig:token_perturbation}
    \end{minipage}
  \end{tabular}
  \vspace{-10pt}
\end{figure*}

\begin{figure*}[t]
    \centering
    \includegraphics[width=\linewidth]{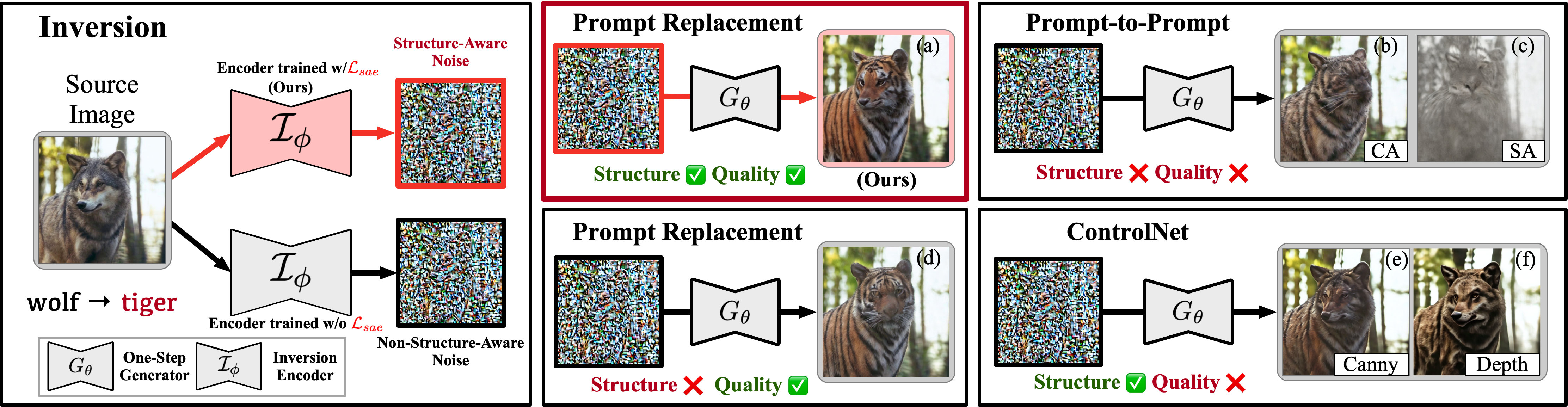}
    \caption{\textbf{Comparison of structure-preserving methods for one-step editing.} (a) Our method, trained with the proposed $\mathcal{L}_\mathrm{sae}$, maintains structural integrity. (d) Without $\mathcal{L}_\mathrm{sae}$, our model fails, leading to structural collapse. (b, c, e, f) Similarly, existing control methods often produce degraded results, demonstrating their unsuitability for one-step generation. SA and CA denote the replacement of self-attention and cross-attention, respectively.}
    \label{fig:attentioninjection}
    \vspace{-10pt}
\end{figure*}

To generate such a dataset $DB = \{\, (z^0, z^0_{\mathrm{edit}}, c, c_{\mathrm{edit}}) \,\}$, a natural approach would be to apply an off-the-shelf image editor to each source image.
However, such editors typically rely on different generative backbones, and their outputs may fall outside the reachable manifold of the target one-step generator $G_\theta$.
Training $\mathcal{I}_\phi$ on these distributionally mismatched targets
leads to inconsistent gradients, as the encoder is asked to predict
latents for images that $G_\theta$ could never have produced in the
first place.
Table~\ref{tab:off-the-shelf} confirms this: an encoder trained on
off-the-shelf\footnote{\label{fn:offtheshelf}We use
Google Gemini's Nano Banana as a representative off-the-shelf image
editor with default recommended settings.} editing pairs yields notably lower reconstruction quality than ours across all metrics, indicating that the mismatched supervision corrupts the encoder's overall latent representation.
We also report DDIM 50-step inversion~\cite{song2020ddim} as a
baseline, which performs worst due to the fundamental
inversion--generation mismatch discussed in
Section~\ref{sec:intro}.

To avoid this mismatch, we propose a simple yet effective alternative termed \textit{prompt perturbation}, which ensures that every target image is generator-aligned, \ie, lies within $G_\theta$'s output distribution by construction.
Concretely, we first sample a text prompt from a corpus (LAION~\cite{schuhmann2021laion}, JourneyDB~\cite{sun2023journeydb}) and a random noise vector $z^T \sim \mathcal{N}(0,I)$. Let $c = E(y)$ be the text embedding of the sampled prompt $y$. We then construct a perturbed embedding as:
\begin{align}
c_{\mathrm{edit}} = c + \lambda_{\mathrm{noise}} \cdot \eta, 
\quad \eta \sim \mathcal{N}(0,I),
\end{align}
where $\lambda_{\mathrm{noise}}$ controls the perturbation intensity.
Because both the source image $G_\theta(z^T, c)$ and the edited image $G_\theta(z^T, c_{\mathrm{edit}})$ are generated by the same frozen generator from the same initial noise, the resulting pairs are sampled from $G_\theta$'s reachable output distribution and tend to share a similar structural layout while primarily differing in appearance.
This generator-aligned supervision provides a well-posed training signal for $\mathcal{I}_\phi$, which must predict latents that $G_\theta$ can faithfully decode.

We empirically study the effect of $\lambda_{\mathrm{noise}}$ on both structure and semantics (Figure~\ref{fig:token_perturbation}). Starting from the base prompt ``a photo of a dog'', we vary $\lambda_{\mathrm{noise}} \in \{0.0, 0.1, \dots, 0.5\}$ and, for each
setting, generate 100 image pairs using the same initial noise $z^T$. For each pair, we measure (i)~a structural distance using DINO~\cite{caron2021emerging}, and (ii)~a semantic distance using CLIP~\cite{radford2021clip}. As $\lambda_{\mathrm{noise}}$ increases, the semantic distance grows, while for appropriately small noise (\eg, $\lambda_{\mathrm{noise}} = 0.1$) the structural distance remains low and stable, confirming that small perturbations preserve layout while inducing meaningful appearance changes. Guided by this analysis, we set $\lambda_{\mathrm{noise}} = 0.1$ in all experiments (see Supplementary~B and G for training results of other parameters).

Given $(c, c_{\mathrm{edit}})$ and a shared $z^T$, we obtain the reconstruction and editing targets as
\begin{align}
\underbrace{z^0 = G_\theta(z^T, c)}_{\text{Reconstruction Target}},\quad
\underbrace{z^0_{\mathrm{edit}} = G_\theta(z^T, c_{\mathrm{edit}})}_{\text{Editing Target}}.
\end{align}
This process yields a large dataset in which most pairs exhibit strong structural similarity but differ in appearance, effectively simulating the desired editing behavior.


\subsubsection{Training with Structure-Aware Editing Loss.}

The dataset provides pairs that share structure but differ in semantics, each pair generated from the same noise $z^T$ under the original and perturbed prompts.
We translate this into a training objective that forces the inversion encoder to produce a latent $\hat{z}^T = \mathcal{I}_\phi(z^0, c)$ serving both reconstruction and structure-preserving editing.
Our loss is:
\begin{align}
\label{eq:total_loss}
\mathcal{L}_{\mathrm{total}}
=
\underbrace{\bigl\|
   z^0
 - G_{\theta}(\mathcal{I}_{\phi}(z^0, c),\; c)
\bigr\|_2^2}_{\mathcal{L}_{\mathrm{mse}}}
\;+\;
\lambda_{\mathrm{sae}}\;
\underbrace{\bigl\|
   z^0_\mathrm{edit}
 - G_{\theta}(\mathcal{I}_{\phi}(z^0, c),\; c_{\mathrm{edit}})
\bigr\|_2^2}_{\mathcal{L}_{\mathrm{sae}}}\;,
\end{align}
where $\lambda_{\mathrm{sae}}$ is the weight parameter (set to 1.0), $z^0 = G_{\theta}(z^T, c)$, and $z^0_{\mathrm{edit}} = G_{\theta}(z^T, c_{\mathrm{edit}})$.

The first term, $\mathcal{L}_{\mathrm{mse}}$, ensures that the inverted latent allows $G_\theta$ to accurately reconstruct the original image $z^0$.
Reconstruction alone, however, does not guarantee that the same latent will preserve structure when the prompt changes.
The second term, the Structure-Aware Editing (SAE) loss $\mathcal{L}_{\mathrm{sae}}$, directly addresses this: it requires that the same inverted latent $\hat{z}^T$, when decoded with the perturbed prompt $c_{\mathrm{edit}}$, produces an output matching the pre-generated target $z^0_{\mathrm{edit}}$.
Because $z^0_{\mathrm{edit}}$ was generated from the same noise $z^T$ as $z^0$, it shares the source structure by construction, so matching it teaches the encoder to produce latents whose structural content is invariant to prompt changes.
Figure~\ref{fig:attentioninjection} validates this design: without $\mathcal{L}_{\mathrm{sae}}$, the model suffers from structural collapse (d), while existing control methods such as attention injection (b, c, e, f) similarly fail, confirming that structure preservation must come from the latent itself rather than from external control during generation.

For implementation, we initialize the inversion encoder $\mathcal{I}_{\phi}$ with the pre-trained one-step model $G_\theta$, leveraging components from its U-Net. Detailed training settings and visual and statistical analyses of the latents $\hat{z}^T$ produced by $\mathcal{I}_{\phi}$ are provided in Supplementary~C.

\begin{figure}[t]
    \centering
    \includegraphics[width=1\linewidth]{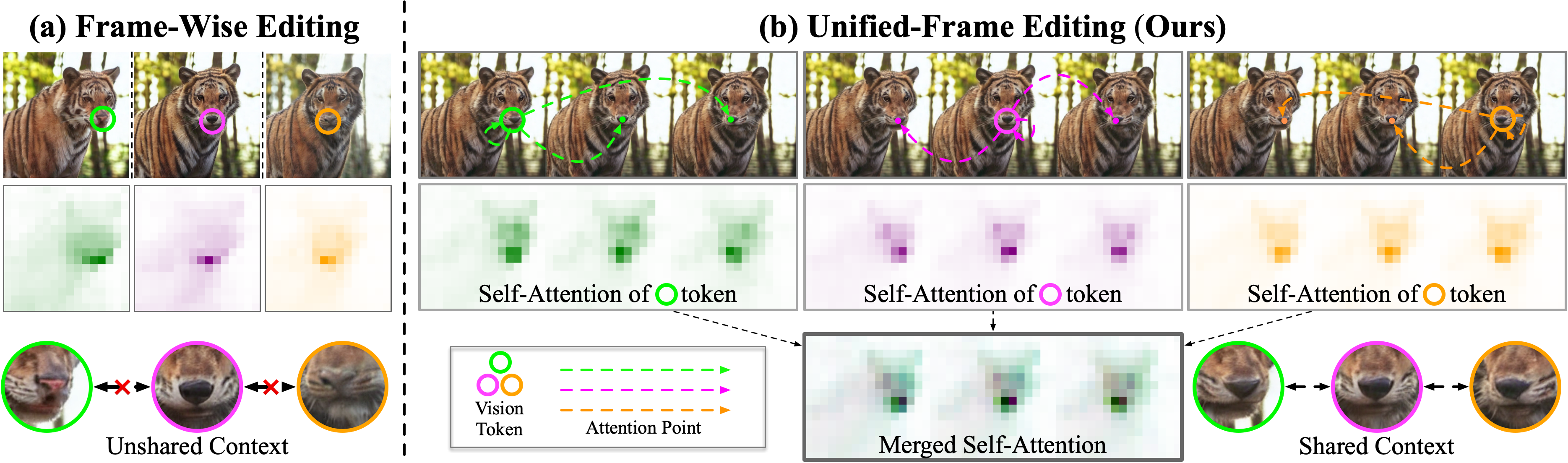}
    \caption{\textbf{Mechanism of Unified-Frame Editing for Temporal Consistency.} (a) Frame-Wise Editing: Processing frames individually restricts attention to within each frame, preventing information sharing and causing temporal incoherence. (b) Unified-Frame Editing (UFE): Our method concatenates all frame latents, allowing the attention mechanism to operate globally. This enables features to be matched and aligned across all frames, enforcing temporal consistency throughout the edit.}
    \label{fig:UFE-Attention}
\end{figure}

\subsection{Unified-Frame Editing}

\label{sec:3.2}
While the inversion encoder from Section~\ref{sec:3.1} is effective for single images, applying it
frame-wise to video does not preserve temporal consistency.
Most prior video editing methods enforce consistency by injecting guidance signals into intermediate
features or attention maps during multi-step
denoising~\cite{cong2024flattenopticalflowguidedattention,geyer2023tokenflow,yang2024frescospatialtemporalcorrespondencezeroshot,kara2024rave,wang2024cove,liu2024video,wu2023tuneavideooneshottuningimage}.
However, as discussed in Section~\ref{sec:3.1}, such iterative guidance mechanisms are inapplicable in the one-step regime (Figure~\ref{fig:attentioninjection}).

This motivates a fundamentally different strategy: rather than manipulating internal attention, we
encourage cross-frame communication by modifying only the \emph{input} latent representation.
We propose \textbf{Unified-Frame Editing (UFE)}, an input-level latent unification scheme in which
inverted latents from multiple frames are spatially concatenated and processed simultaneously in a
single forward pass (Figure~\ref{fig:mainfigure}(b)).

Concretely, given a video sequence of $K$ frames
$\mathcal{V}_K^0 = (z^0_0, \dots, z^0_{K-1})$,
each frame is independently inverted via our encoder:
\begin{align}
\hat{z}^T_k = \mathcal{I}_{\phi}(z^0_k,\, c),
\quad k = 0, \dots, K-1.
\end{align}
We denote the set of inverted latents as $\hat{\mathcal{V}}_K^T = (\hat{z}^T_0, \dots, \hat{z}^T_{K-1})$.
Let $\hat{z}^T_k \in \mathbb{R}^{C \times H \times W}$ denote the inverted latent for the $k$-th frame. The unified latent map $Z^T_{\mathrm{UFE}}$ is formed by concatenating all $K$ latents along the width dimension:
\begin{align}
Z^T_{\mathrm{UFE}} = \mathrm{Concat}(\hat{z}^T_0, \hat{z}^T_1, \dots, \hat{z}^T_{K-1}, \text{axis}=W).
\end{align}
The resulting map $Z^T_{\mathrm{UFE}}$ has dimensions $\mathbb{R}^{C \times H \times (W \cdot K)}$ and is passed to the one-step generator $G_\theta(\cdot)$ with the editing prompt $c_{\mathrm{edit}}$.
This is feasible because the U-Net operates on spatial features and naturally accepts inputs of
arbitrary size~\cite{long2015fully, ronneberger2015u, rombach2022high}.
Crucially, this design allows the generator's self-attention to function globally across all frames,
leveraging its ability to align corresponding features~\cite{tang2023emergent} and thereby enforce
temporal consistency without any external guidance module.
Figure~\ref{fig:UFE-Attention} visualizes this effect: unlike frame-wise editing where information
remains isolated, a token representing a specific feature (\eg, the tiger's nose) in one frame can
directly attend to its counterpart in other frames, ensuring that edits are applied consistently
across the entire sequence.

\subsubsection{Sliding Window with Anchor Frame.}
While UFE is effective, its VRAM requirement and inference time scale with the number of frames (Supplementary H), making it computationally challenging for long videos. To address this, we extend UFE with a sliding window mechanism guided by a global anchor frame. Figure~\ref{fig:mainfigure}(b) illustrates the pipeline of our method. The core idea is to process the video in manageable chunks while ensuring both local and global temporal consistency. The sliding window provides local coherence, while the anchor frame enforces a consistent global style and content reference across all windows. First, we identify a representative anchor frame by selecting the medoid in the DINO~\cite{caron2021emerging} feature space. Its inverted latent, $\hat{z}^T_A$, serves as a global anchor. We validate this choice against alternative strategies, including the use of multiple anchors for videos with scene transitions or dynamic content changes, in Supplementary D.

We then process the video using an overlapping sliding window approach. For each window of $w$ inverted latents, we prepend the anchor latent $\hat{z}^T_A$ before feeding the concatenated map to the generator $G_\theta$. After generation, the output corresponding to the anchor is discarded, and we extract only the central $s$ frames from the window's output. These segments are then seamlessly stitched together to form the final edited video. This anchor-based sliding window strategy effectively prevents temporal drift by ensuring that each segment is coherent both locally and globally. We set the window size as $w=7$ and a stride of $s=5$ through ablation (Supplementary E). A detailed, step-by-step formulation of this process is provided in Supplementary F.

\vspace{-10pt}
\section{Experiments}

\begin{table}[t]
\centering
\caption{Performance comparison on short (20 frames) videos.}
\vspace{-10pt}
\setlength{\tabcolsep}{2pt}
\renewcommand{\arraystretch}{0.90}
\resizebox{\linewidth}{!}{%
\begin{tabular}{l l *{6}{c} >{\columncolor{gray!15}}c >{\columncolor{yellow!15}}c}
\toprule
\textbf{Framework} & \textbf{Method}
& SC & BC & TF & MS & AQ & IQ
& \cellcolor{gray!15}BQS & \cellcolor{yellow!15}FPS \\
\midrule

& FLATTEN$^{\ddagger}$
& 0.965 & 0.970 & 0.964 & 0.972 & 0.625 & 0.639 & 0.611 & 0.072 \\

& TokenFlow$^{\ddagger}$
& \textbf{0.983} & 0.976 & 0.985 & 0.991 & 0.668 & 0.680 & 0.663 & 0.075 \\

Multi-to-Multi & FRESCO$^{\ddagger}$
& 0.978 & 0.974 & 0.973 & 0.991 & 0.649 & \textbf{0.729} & 0.674 & 0.078 \\

& RAVE$^{\ddagger}$
& 0.982 & 0.976 & 0.975 & 0.986 & 0.637 & 0.695 & 0.653 & 0.091 \\

& COVE$^{\ddagger}$
& \textbf{0.983} & 0.976 & 0.984 & 0.989 & 0.645 & 0.655 & 0.639 & 0.061 \\
\cmidrule(r){1-10}

& Prompt Replacement$^{\dagger}$
& 0.921 & 0.946 & 0.948 & 0.979 & 0.593 & 0.562 & 0.548 & 4.732 \\

& Prompt-to-Prompt$^{\dagger}$
& 0.915 & 0.945 & 0.965 & 0.978 & 0.587 & 0.566 & 0.548 & 0.898 \\

Multi-to-One & ControlNet (Depth)$^{\dagger}$
& 0.968 & 0.961 & 0.970 & 0.984 & 0.658 & 0.673 & 0.646 & 0.578 \\

& ControlNet (Canny)$^{\dagger}$
& 0.960 & 0.969 & 0.966 & 0.983 & 0.582 & 0.662 & 0.603 & 0.578 \\

& Plug-and-Play$^{\dagger}$
& 0.953 & 0.971 & \textbf{0.995} & \textbf{0.995} & 0.486 & 0.220 & 0.345 & 0.398 \\
\cmidrule(r){1-10}

\textbf{One-to-One} & \textbf{\Ours\ (Ours)}$^{\dagger}$
& \textbf{0.983} & \textbf{0.977} & 0.978 & 0.991 & \textbf{0.678} & 0.703 & \textbf{0.679} & \textbf{15.625} \\
\bottomrule
\end{tabular}
}
\vspace{-10pt}
\begin{flushleft}
\scriptsize
$^{\ddagger}$~Multi-Step Diffusion Model (SD1.5).\quad
$^{\dagger}$~One-Step Diffusion Model (DMD2).
\end{flushleft}
\vspace{-10pt}
\label{tab:video_results_20}
\end{table}

\begin{table}[t]
\centering
\caption{Performance comparison on long (90 frames) videos.}
\vspace{-10pt}
\setlength{\tabcolsep}{2pt}
\renewcommand{\arraystretch}{0.90}
\resizebox{\linewidth}{!}{%
\begin{tabular}{l l *{6}{c} >{\columncolor{gray!15}}c >{\columncolor{yellow!15}}c}
\toprule
\textbf{Framework} & \textbf{Method}
& SC & BC & TF & MS & AQ & IQ
& \cellcolor{gray!15}BQS & \cellcolor{yellow!15}FPS \\
\midrule

& FLATTEN$^{\ddagger}$
& 0.933 & 0.960 & 0.965 & 0.975 & 0.648 & 0.636 & 0.615 & 0.052 \\

& TokenFlow$^{\ddagger}$
& \textbf{0.974} & \textbf{0.975} & 0.986 & 0.990 & 0.622 & 0.668 & 0.633 & 0.084 \\

Multi-to-Multi & FRESCO$^{\ddagger}$
& 0.956 & 0.968 & 0.987 & 0.992 & 0.640 & 0.702 & 0.655 & 0.087 \\

& RAVE$^{\ddagger}$
& 0.962 & 0.963 & 0.978 & 0.983 & 0.665 & 0.692 & 0.659 & 0.102 \\

& COVE$^{\ddagger}$
& 0.955 & 0.964 & 0.985 & 0.988 & 0.651 & 0.650 & 0.633 & 0.041 \\
\cmidrule(r){1-10}

& Prompt Replacement$^{\dagger}$
& 0.841 & 0.912 & 0.970 & 0.975 & 0.561 & 0.528 & 0.503 & 4.732 \\

& Prompt-to-Prompt$^{\dagger}$
& 0.900 & 0.956 & 0.969 & 0.974 & 0.423 & 0.267 & 0.328 & 0.898 \\

Multi-to-One & ControlNet (Depth)$^{\dagger}$
& 0.945 & 0.944 & 0.971 & 0.978 & 0.626 & 0.652 & 0.613 & 0.578 \\

& ControlNet (Canny)$^{\dagger}$
& 0.934 & 0.952 & 0.967 & 0.976 & 0.590 & 0.660 & 0.598 & 0.578 \\

& Plug-and-Play$^{\dagger}$
& 0.841 & 0.912 & \textbf{0.994} & \textbf{0.995} & 0.561 & 0.528 & 0.509 & 0.398 \\
\cmidrule(r){1-10}

\textbf{One-to-One} & \textbf{\Ours\ (Ours)}$^{\dagger}$
& 0.958 & 0.965 & 0.978 & 0.989 & \textbf{0.670} & \textbf{0.723} & \textbf{0.677} & \textbf{15.793} \\
\bottomrule
\end{tabular}
}
\vspace{-10pt}
\begin{flushleft}
\scriptsize
$^{\ddagger}$~Multi-Step Diffusion Model (SD1.5).\quad
$^{\dagger}$~One-Step Diffusion Model (DMD2).
\end{flushleft}
\vspace{-10pt}
\label{tab:video_results_90}
\end{table}
\begin{figure}[t]
    \centering
    \includegraphics[width=0.95\linewidth]{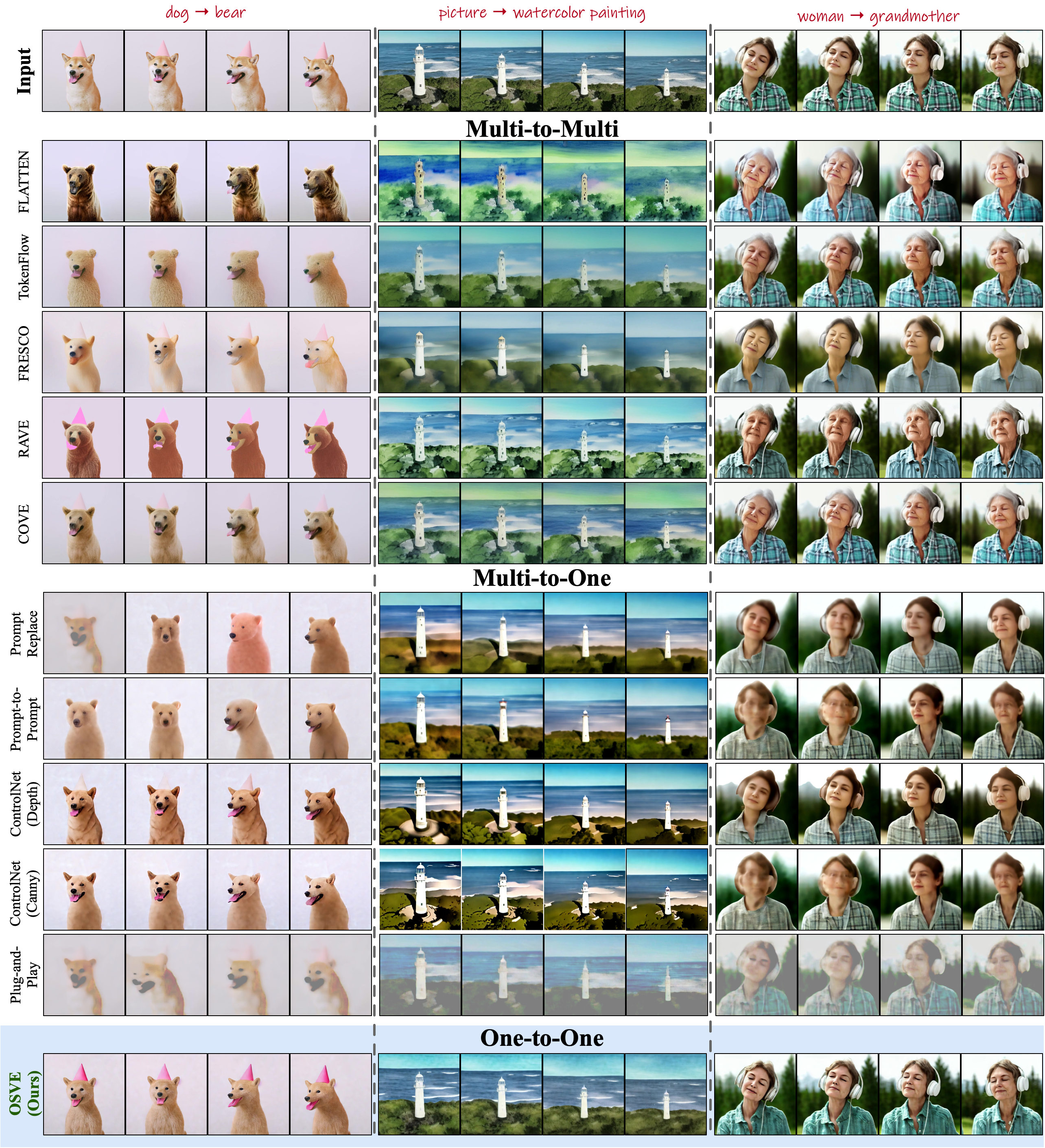}
    \caption{Qualitative comparison with baseline methods.}
    \label{fig:qualitative1}
\end{figure}
\vspace{-10pt}
\begin{figure}[t]
    \centering
    \includegraphics[width=0.95\linewidth]{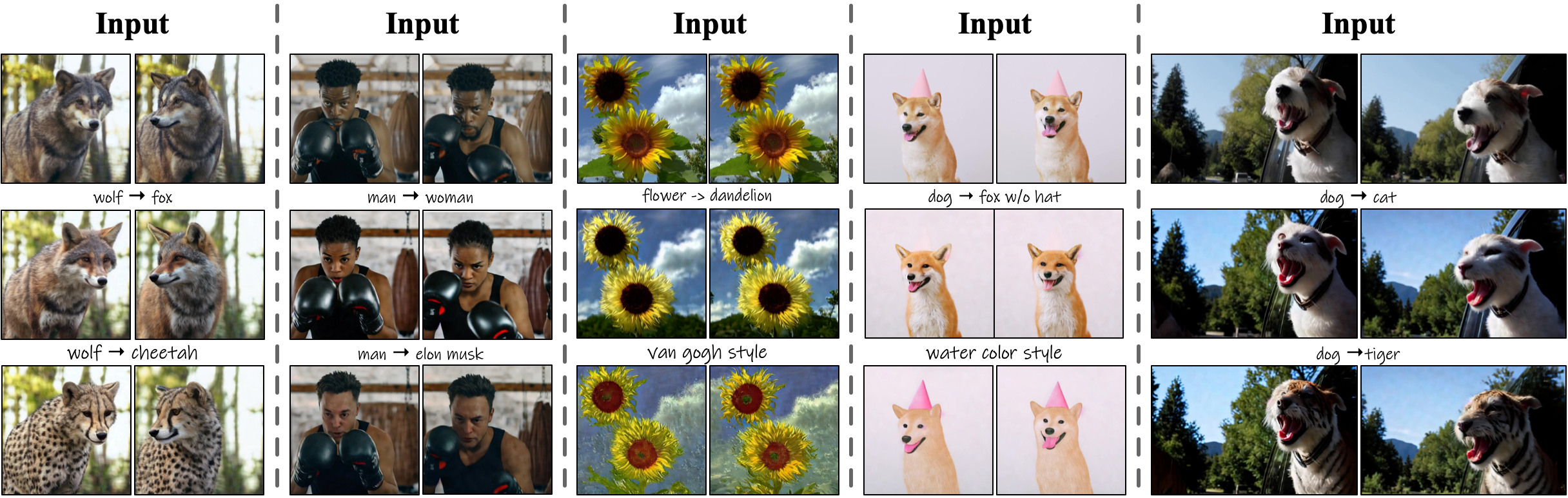}
    \caption{Qualitative results with OSVE across various editing tasks.}
    \label{fig:qualitative2}
\vspace{-10pt}
\end{figure}
\subsection{Experimental Setup}
We evaluate our framework across three video editing settings. The first, \textbf{Multi-to-Multi}, uses multi-step diffusion models for both inversion and editing. We compare against five baselines: FLATTEN~\cite{cong2024flattenopticalflowguidedattention}, TokenFlow~\cite{qu2024tokenflowunifiedimagetokenizer}, FRESCO~\cite{yang2024frescospatialtemporalcorrespondencezeroshot}, RAVE~\cite{kara2024rave}, and COVE~\cite{wang2024cove}, all built on Stable Diffusion 1.5~\cite{rombach2022high}. The second, \textbf{Multi-to-One}, performs video inversion via multi-step DDIM while editing in a single step. As no prior work exists for this configuration, we construct baselines by adapting image editing techniques: Prompt Replacement, Prompt-to-Prompt~\cite{hertz2023prompttoprompt}, ControlNet~\cite{zhang2023adding} with depth and Canny guidance, and Plug-and-Play~\cite{tumanyan2023plug}. The third is our proposed \textbf{One-to-One} framework, which employs the pre-trained encoder for single-step inversion and edits via Prompt Replacement. For the Multi-to-One and One-to-One settings, we use DMD2~\cite{yin2024improved} as the one-step diffusion backbone. Further details are provided in Supplementary~A.

For evaluation, we curated 60 videos from open platforms such as Pixabay and prior works, comprising 51 short videos (20 frames) and 9 long videos (90 frames). We designed five editing prompts per video---three for local edits (\eg, object change, deletion) and two for global edits (\eg, style transfer, background modification)---resulting in 255 short-video and 45 long-video evaluation pairs. Short videos are processed as a single window, while long videos use a sliding window of 7 frames with a stride of 5. All experiments were conducted on a single NVIDIA RTX 6000 Ada GPU.
\vspace{-10pt}
\subsection{Quantitative Comparison}
We adopt VBench~\cite{huang2024vbench} metrics for quantitative evaluation, organized into two categories. \textbf{Temporal consistency} metrics include: Subject Consistency (SC), measuring subject identity preservation via DINO~\cite{caron2021emerging} features; Background Consistency (BC), assessing background stability via CLIP~\cite{radford2021clip} features; Temporal Flickering (TF), where higher indicates less flickers; and Motion Smoothness (MS), evaluating motion fluidity using a pre-trained video interpolation model~\cite{li2023amt}. \textbf{Per-frame quality} metrics include: Aesthetic Quality (AQ), scored by the LAION aesthetic predictor~\cite{laion2022aesthetic}, and Imaging Quality (IQ), quantifying perceptual artifacts using MUSIQ~\cite{ke2021musiq}. We also introduce the Balanced Quality Score (BQS): $\mathrm{BQS}=\left(\frac{\mathrm{SC}+\mathrm{BC}+\mathrm{TF}+\mathrm{MS}}{4}\right)\times\left(\frac{\mathrm{AQ}+\mathrm{IQ}}{2}\right)$, which balances both dimensions multiplicatively, penalizing methods that excel in only one.

Tables~\ref{tab:video_results_20} and~\ref{tab:video_results_90} present the results for short and long videos, respectively. Our method, OSVE, achieves the highest BQS in both settings, demonstrating strong temporal consistency and per-frame quality simultaneously. In terms of efficiency, OSVE is approximately $171\times$ faster for short videos and $155\times$ faster for long videos compared to RAVE, the fastest Multi-to-Multi baseline. See Supplementary~A for the speed comparison setup.
\vspace{-10pt}
\subsection{Qualitative Comparison}
Figure~\ref{fig:qualitative1} provides a qualitative comparison. OSVE successfully modifies objects while preserving their underlying structure, whereas the Multi-to-One baselines (rows~7--11) frequently exhibit structural collapse or visual degradation. Figure~\ref{fig:qualitative2} illustrates the versatility of OSVE across diverse editing tasks, including object replacement (columns~1, 3, and~5), human identity modification (column~2), artistic style transfer (columns~3 and~4), and prompt-guided object removal (column~4).
\vspace{-10pt}
\subsection{User Study}
We conducted a user study with 26 participants. Each participant viewed six editing methods applied to three randomly sampled videos from our datasets and rated each result on Temporal Consistency (TC) and Visual Quality (VQ) using a 5-point Likert scale. As shown in Table~\ref{tab:user_study}, OSVE achieves the highest scores on both dimensions, outperforming all baselines.
\vspace{-10pt}
\subsection{Ablation Study}
\subsubsection{Inversion Encoder.}
Table~\ref{tab:encoder_ablation} validates the Structure-Aware Editing (SAE) Loss on the PIE-Bench~\cite{ju2024pnp}, using structure distance~\cite{tumanyan2023plug} and CLIP scores~\cite{radford2021clip}. While SAE Loss alone improves structure preservation, combining it with the reconstruction loss ($\mathcal{L}_\mathrm{mse}$) yields the best performance in both structural integrity and editability, indicating a synergistic effect between the two objectives.
\vspace{-10pt}
\subsubsection{Unified Frame Editing.}
Table~\ref{tab:ufe_abl} presents the ablation study of Unified-Frame Editing (UFE). Both the sliding window and anchor-frame strategies individually improve temporal consistency (SC, BC, TF, MS), and their combination yields the strongest results. We observe a marginal decrease in AQ when applying UFE, likely due to processing the expanded latent representation. However, this minor trade-off is perceptually negligible compared to the substantial reduction in flickering and inconsistencies.

\begin{table}[t]
\centering
\small
\setlength{\tabcolsep}{3pt}

\begin{minipage}[t]{0.40\linewidth}
\centering
\renewcommand{\arraystretch}{0.85}
\caption{User study results. TC: Temporal Consistency, VQ: Visual Quality (higher is better).}
\begin{tabularx}{\linewidth}{l >{\centering\arraybackslash}X >{\centering\arraybackslash}X}
\toprule
\textbf{Method} & \textbf{TC} & \textbf{VQ} \\
\midrule
FLATTEN              & 2.59             & 2.56             \\
TokenFlow            & \underline{3.68} & 3.08             \\
FRESCO               & 3.02             & 2.76             \\
RAVE                 & 2.74             & 2.70             \\
COVE                 & 3.59             & \underline{3.41} \\
\midrule
\rowcolor{gray!15}
\textbf{OSVE (Ours)} & \textbf{3.85}    & \textbf{3.65}    \\
\bottomrule
\end{tabularx}
\label{tab:user_study}
\end{minipage}
\hfill
\begin{minipage}[t]{0.56\linewidth}
\centering
\renewcommand{\arraystretch}{1.1}
\caption{Ablation study on encoder training. We evaluate the effect of each loss component on PIE-Bench.}
\begin{tabularx}{\linewidth}{cc >{\centering\arraybackslash}X >{\centering\arraybackslash}X >{\centering\arraybackslash}X}
\toprule
$\mathcal{L}_{\text{mse}}$ &
$\mathcal{L}_{\text{sae}}$ &
\makecell[c]{Struct.\\Dist. ($\downarrow$)} &
\makecell[c]{CLIP\\Whole ($\uparrow$)} &
\makecell[c]{CLIP\\Edit ($\uparrow$)} \\
\midrule
$\checkmark$ & --           & 0.087          & 21.797          & 19.884          \\
--           & $\checkmark$ & 0.074          & \textbf{22.349} & 19.863          \\
\rowcolor{gray!30}
$\checkmark$ & $\checkmark$ & \textbf{0.064} & 22.329          & \textbf{20.416} \\
\bottomrule
\end{tabularx}
\label{tab:encoder_ablation}
\end{minipage}

\end{table}

\begin{table}[t]
\centering
\small
\setlength{\tabcolsep}{4pt}
\renewcommand{\arraystretch}{1.30}
\fontsize{7.6pt}{8.0pt}\selectfont
\caption{Ablation study of UFE on 90-frame videos.}
\vspace{-10pt}
\begin{tabularx}{\linewidth}{cc *{7}{>{\centering\arraybackslash}X}}
\toprule
\makecell[c]{Sliding\\window} & Anchor & {SC} & {BC} & {TF} & {MS} & {AQ} & {IQ} & {BQS}\\
\midrule
--           & --           & {0.931}          & {0.948}          & {0.972}        & {0.980}          & {\textbf{0.680}} & {0.722}          & {0.671}          \\
$\checkmark$ & --           & {0.954}          & {0.963}          & {0.977}        & {0.988}          & {0.671}          & {0.723}          & {0.676}          \\
--           & $\checkmark$ & {0.943}          & {0.950}          & {0.976}        & {0.985}          & {0.672}          & {0.722}          & {0.669}          \\
\rowcolor{gray!30}
$\checkmark$ & $\checkmark$ & {\textbf{0.958}} & {\textbf{0.965}} & \textbf{0.978} & {\textbf{0.989}} & {0.670}          & {\textbf{0.723}} & {\textbf{0.677}} \\
\bottomrule
\end{tabularx}
\label{tab:ufe_abl}
\vspace{-10pt}
\end{table}

\section{Discussion}
\vspace{-4pt}
We build OSVE on a one-step T2I backbone because current one-step T2V generators, despite recent progress in T2V distillation~\cite{zheng2025large,yang2025causvid,lin2025seaweedapt}, still suffer from severe temporal artifacts and blurriness that make them unsuitable as a reliable editing backbone (see Supplementary~G).
In contrast, mature one-step T2I backbones have been reported to
even surpass their multi-step teachers in generation
quality~\cite{yin2024improved}, providing a clean testbed to evaluate
our core contributions independently of backbone limitations.

The generative landscape is rapidly expanding toward T2V, T2AV, and
other multimodal architectures, and one-step distillation of these
models is an active area of
research~\cite{lin2025seaweedapt,lin2024animatedifflightning,
li2024t2vturbo,zhang2024sfv,ding2025dollar}.
As their fidelity improves, extending one-step editing to these
backbones will become both feasible and important.
We believe the core principles behind OSVE, namely encoding structure into the initial latent and enforcing cross-frame consistency at the input level, are not specific to T2I and can generalize across modalities.
\vspace{-10pt}
\section{Conclusion}
\vspace{-4pt}
We presented \textbf{OSVE}, a framework that advances text-guided video editing by introducing a novel One-to-One pipeline. OSVE unlocks the potential of one-step diffusion models for this task, overcoming the speed limitations of conventional multi-step methods. Our two core contributions, a fast learnable inversion encoder trained with \textbf{SAE} loss to preserve structure and the \textbf{UFE} mechanism to enforce temporal coherence, resolve the key bottlenecks that previously hindered this approach. Our experiments show that OSVE achieves competitive quality and consistency at a fraction of the computational cost, making real-time video editing practical and accessible.

\section*{Acknowledgements}
This work was supported by the National Research Foundation of Korea (NRF) grant funded by the Korea government (MSIT) (RS-2026-25480253), and in part by Korea Planning \& Evaluation Institute of Industrial Technology (KEIT) grant funded by the Korea government (MOTIE) (RS-2024-00444344), and by the ``Advanced GPU Utilization Support Program'' funded by the Government of the Republic of Korea (Ministry of Science and ICT).
\newcounter{tocmain}
\newcounter{tocsub}[tocmain]

\renewcommand{\thetocmain}{\Alph{tocmain}}
\renewcommand{\thetocsub}{\thetocmain.\arabic{tocsub}}

\newcommand{\TOCMain}[2]{%
\refstepcounter{tocmain}%
\setcounter{tocsub}{0}%
\noindent\hyperref[#1]{\textbf{\thetocmain. #2}} \dotfill \pageref{#1}\par
}

\newcommand{\TOCSub}[2]{%
\refstepcounter{tocsub}%
\hspace*{1.5em}\hyperref[#1]{\thetocsub\ #2} \dotfill \pageref{#1}\par
}

\newcommand{\TOCNew}{\addvspace{0.3em}}
\section*{Table of Contents}
\vspace{0.5em}

\TOCMain{sec:A}{Experimental Details}
\TOCSub{sec:A1}{Multi-to-Multi}
\TOCSub{sec:A2}{Multi-to-One}
\TOCSub{sec:A3}{One-to-One}
\TOCSub{sec:A4}{Speed Comparison}
\TOCNew

\TOCMain{sec:B}{Ablation on Prompt Perturbation}
\TOCNew

\TOCMain{sec:C}{Encoder Training and Analysis of Inverted Latents}
\TOCSub{sec:C1}{Training Details}
\TOCSub{sec:C2}{Visualization and Analysis}
\TOCNew

\TOCMain{sec:D}{Anchor Ablations}
\TOCSub{sec:D1}{Anchor selection strategies ablation}
\TOCSub{sec:D2}{Multi-Anchor Ablation}
\TOCNew

\TOCMain{sec:E}{Ablation on Sliding Window and Anchor}
\TOCNew

\TOCMain{sec:F}{Unified-Frame Editing}
\TOCNew

\TOCMain{sec:G}{Additional Analysis}
\TOCSub{sec:G1}{Detailed Ablation on $\lambda_{\mathrm{noise}}$}
\TOCSub{sec:G2}{Discussion on One-Step T2V Generator Quality}
\TOCNew

\TOCMain{sec:H}{GPU Memory Analysis}
\TOCMain{sec:I}{User Study}

\vspace{1em}
\hrule
\vspace{1em}

\setcounter{section}{0}
\renewcommand{\thesection}{\Alph{section}}  

\setcounter{figure}{0}
\renewcommand{\thefigure}{\Alph{figure}}    

\setcounter{table}{0}
\renewcommand{\thetable}{\Alph{table}}      

\section{Experimental Details}\label{sec:A}

In this section, we provide comprehensive implementation details for all methods evaluated in our paper. All training and inference experiments are conducted on a single NVIDIA RTX 6000 Ada graphics card (48\,GB VRAM). Unless stated otherwise, we follow the hyperparameter settings reported in each respective paper. For cases where specific hyperparameters are not explicitly declared, we adopt the default values provided in the official open-source implementations.

To ensure fair comparison across all methods, we keep the input video resolution and frame count consistent throughout our experiments. All input videos are resized to $512 \times 512$ resolution. We use the same set of source videos and editing prompts for every method under the same experimental setting.

\subsection{\textbf{Multi-to-Multi}}\label{sec:A1}

The \textbf{Multi-to-Multi} setting refers to the conventional video editing pipeline where both the inversion and sampling processes require multiple diffusion steps. In this setting, we adopt Stable Diffusion v1-5\footnote{\url{https://huggingface.co/runwayml/stable-diffusion-v1-5}} as the pre-trained multi-step diffusion model for all methods. For the inversion process, we uniformly apply the DDIM Inversion scheduler~\cite{song2020ddim} with 50 inversion steps across all methods, which reconstructs the latent noise representation from each source video frame. We describe the specific configurations for each baseline method below.

\vspace{0.5em}
\noindent\textbf{FLATTEN}~\cite{cong2024flattenopticalflowguidedattention}.
We use the official PyTorch implementation of FLATTEN\footnote{\url{https://github.com/yrcong/flatten}}, which leverages optical flow-guided attention to improve temporal consistency in video editing. Following the settings from the original paper, we set the classifier-free guidance scale to 20 and use 50 DDIM sampling steps for the denoising process. FLATTEN modifies the attention mechanism in the U-Net by incorporating optical flow information, which guides the attention maps to be temporally coherent across frames. We use the default optical flow estimation model provided in the official repository.

\vspace{0.5em}
\noindent\textbf{TokenFlow}~\cite{geyer2023tokenflow}.
We use the official PyTorch implementation of TokenFlow\footnote{\url{https://github.com/omerbt/TokenFlow}}. TokenFlow enforces temporal consistency by propagating diffusion features across frames based on inter-frame correspondences. Following the paper's methodology, we utilize the Plug-and-Play (PnP) editing framework~\cite{tumanyan2023plug} as the underlying editing mechanism. The classifier-free guidance scale is set to 7.5, and we use 50 DDIM sampling steps. TokenFlow first identifies keyframes, computes nearest-neighbor correspondences between them, and then propagates the edited tokens to all other frames to maintain visual coherence.

\vspace{0.5em}
\noindent\textbf{FRESCO}~\cite{yang2024frescospatialtemporalcorrespondencezeroshot}.
We use the official PyTorch implementation of FRESCO\footnote{\url{https://github.com/williamyang1991/FRESCO}}, which introduces spatial-temporal correspondences for zero-shot video editing. As recommended in the paper, we employ ControlNet~\cite{zhang2023adding} with HED (Holistically-Nested Edge Detection) conditioning\footnote{\url{https://huggingface.co/lllyasviel/sd-controlnet-hed}} to provide structural guidance during the editing process. We use 20 DDIM sampling steps and set the classifier-free guidance scale to 7.5. We select one keyframe every 8 frames for the editing process. After editing the keyframes, we use EbSynth\footnote{\url{https://github.com/jamriska/ebsynth}}, a patch-based video stylization tool, to propagate the edits from keyframes to all remaining frames and produce the final edited video. All other parameters, including the spatial-temporal attention fusion weights, follow the default settings in the official implementation.

\vspace{0.5em}
\noindent\textbf{RAVE}~\cite{kara2024rave}.
We use the official PyTorch implementation of RAVE\footnote{\url{https://github.com/RehgLab/RAVE}}, which achieves temporally consistent editing by rearranging video frames into a grid layout and processing them simultaneously. Following the settings described in the paper, we employ a ControlNet model conditioned on depth maps\footnote{\url{https://huggingface.co/lllyasviel/sd-controlnet-depth}} to preserve the geometric structure of the source video. The grid size is set to $3 \times 3$ (i.e., 9 frames are arranged in a single grid image), the classifier-free guidance scale is set to 7.5, and we perform 50 DDIM sampling steps. Depth maps are estimated from the source video frames using the default depth estimation model provided in the repository.

\vspace{0.5em}
\noindent\textbf{COVE}~\cite{wang2024cove}.
We use the official PyTorch implementation of COVE\footnote{\url{https://github.com/wangjiangshan0725/COVE}}, which exploits video-level correspondences to guide the diffusion process. Following the settings from the paper, we set the classifier-free guidance scale to 7.5 and the DIFT (Diffusion Features) up-sampling block index to  2, which controls the granularity of the extracted diffusion features used for correspondence computation. The sliding window size for computing inter-frame correspondences is set to 7, the correspondence-guidance scale (which controls the strength of the correspondence constraint during sampling) is set to 3, and the token merging ratio is set to 50\%, meaning half of the tokens are merged based on their correspondence similarities to reduce computational cost while maintaining temporal coherence.

\subsection{\textbf{Multi-to-One}}\label{sec:A2}

The \textbf{Multi-to-One} setting represents a hybrid approach where the inversion process still requires multiple diffusion steps, but the sampling (generation) is performed in a single step using a distilled one-step diffusion model. In this setting, we use DMD2~\cite{yin2024improved} as our pre-trained one-step diffusion model, which is a distribution matching distillation model capable of generating high-quality images in a single forward pass. For the inversion process, we apply the DDIM Inversion scheduler~\cite{song2020ddim} with 50 inversion steps to obtain the latent noise representation from each source frame, identical to the Multi-to-Multi setting. All inference in this setting is performed in a single denoising step using DMD2. Below, we describe the specific configuration of each baseline method adapted to this setting.

\vspace{0.5em}
\noindent\textbf{Prompt Replacement}.
This is the simplest baseline method, which performs editing by directly replacing the source text prompt with the target text prompt during the one-step sampling process. For example, the source prompt ``A photo of a dog'' is replaced with the target prompt ``A photo of a cat,'' while the inverted latent noise is kept unchanged. No additional structural guidance or attention manipulation is applied. This method serves as a lower bound to evaluate the effectiveness of more sophisticated editing techniques.

\vspace{0.5em}
\noindent\textbf{Prompt-to-Prompt}~\cite{hertz2023prompttoprompt}.
We use the official PyTorch implementation of Prompt-to-Prompt\footnote{\url{https://github.com/google/prompt-to-prompt}}, adapted for the one-step sampling setting. Prompt-to-Prompt achieves localized editing by manipulating the cross-attention maps within the U-Net during the denoising process. Specifically, the cross-attention maps corresponding to the source prompt tokens are injected into the denoising process with the target prompt, allowing the model to preserve the spatial layout of the source image while applying the desired edit. In our evaluation, we inject only the cross-attention maps and deliberately exclude the self-attention maps, as we empirically observed that injecting self-attention maps into the one-step generation pipeline leads to significant degradation in output image quality (see Figure~4 in the main paper).

\vspace{0.5em}
\noindent\textbf{ControlNet (Depth)}~\cite{zhang2023adding}.
We employ a pre-trained ControlNet model conditioned on depth maps\footnote{\url{https://huggingface.co/lllyasviel/sd-controlnet-depth}} to provide structural guidance during the one-step sampling process. Depth maps are extracted from each source video frame and fed into the ControlNet branch, which injects depth-aware features into the U-Net to preserve the spatial structure of the original scene. We set the ControlNet conditioning scale (which controls the strength of the structural guidance) to 0.8. This value was chosen to balance between preserving the source structure and allowing sufficient flexibility for the target prompt to take effect.

\vspace{0.5em}
\noindent\textbf{ControlNet (Canny)}~\cite{zhang2023adding}.
Similarly, we use a pre-trained ControlNet model conditioned on Canny edge maps\footnote{\url{https://huggingface.co/lllyasviel/sd-controlnet-canny}} as another structural guidance baseline. Canny edges are detected from each source video frame and used as conditioning input. The ControlNet conditioning scale is set to 0.8, consistent with the depth-conditioned variant. Compared to depth conditioning, Canny edge conditioning provides finer-grained structural constraints based on edge boundaries, which can be advantageous for preserving sharp object boundaries but may also restrict the editing flexibility in some cases.

\vspace{0.5em}
\noindent\textbf{Plug-and-Play}~\cite{tumanyan2023plug}.
We use the official PyTorch implementation of Plug-and-Play\footnote{\url{https://github.com/MichalGeyer/plug-and-play}}, adapted for the one-step generation setting. Plug-and-Play achieves structure-preserving editing by injecting intermediate spatial features from the source image's denoising process into the target image's generation process. Specifically, we extract the spatial features from the convolutional layers of the U-Net during the source image reconstruction and inject them at the corresponding layers during the target image generation, as described in the original paper. This feature injection constrains the spatial layout and structural details of the edited output to align with the source image while allowing semantic changes dictated by the target prompt.

\subsection{\textbf{One-to-One}}\label{sec:A3}

The \textbf{One-to-One} setting represents the most efficient pipeline, where both the inversion and sampling processes are performed in a single step. This eliminates the computational overhead of iterative DDIM inversion entirely, enabling real-time or near-real-time video editing.

\vspace{0.5em}
\noindent\textbf{OSVE (Ours)}.
In this setting, we use our trained OSVE (One-Step Video Editing) encoder to perform single-step inversion. Given a source video frame, the OSVE encoder predicts the corresponding noise latent in a single forward pass, bypassing the need for iterative DDIM inversion. The predicted noise is then directly passed to the DMD2~\cite{yin2024improved} one-step diffusion model for sampling with the target text prompt. Since our inversion encoder is specifically designed and trained to preserve the structural and geometric information of the source image within the predicted noise representation, edits in this setting are performed using only the target text prompt without any additional structural guidance (e.g., ControlNet or attention injection). This design choice results in a highly streamlined editing pipeline that requires only two forward passes per frame: one through the OSVE encoder for inversion and one through the DMD2 model for generation.

\subsection{Speed Comparison}\label{sec:A4}

For a fair and accurate speed comparison across all methods, we measure only the U-Net inference time, excluding the time required for VAE encoding and decoding operations. This is because the VAE computation cost is identical across all methods that share the same base diffusion model, and thus does not contribute to meaningful differences in computational efficiency. The reported time corresponds to the total inversion and inference time aggregated over all frames in a given video clip. To ensure consistency, all operations are executed sequentially rather than asynchronously or in parallel, so that the measured time accurately reflects the per-frame computational cost of each method. We repeat each timing measurement 10 times and report the average to mitigate variance from GPU scheduling and memory caching effects.

\section{Ablation on Prompt Perturbation}\label{sec:B}
\begin{table}[h]
\centering
\begin{tabular}{cccc}
\toprule  
$\lambda_\mathrm{noise}$ & 
\makecell[c]{Structure\\Distance ($\downarrow$)} &  
\makecell[c]{CLIP\\Whole ($\uparrow$)} &
\makecell[c]{CLIP\\Edit ($\uparrow$)} \\
\midrule
 0.0 & 0.087 & 21.797 & 19.884 \\
 \rowcolor{gray!30} 0.1 & \textbf{0.064} & 22.329 & 20.416 \\
0.2 & 0.068 & 22.314 & 20.406  \\
0.3 & 0.075 & 22.449 & 20.422  \\
0.4 & 0.085 & 22.531 & 20.402  \\
0.5 & 0.101 & 22.641 & 20.424  \\
\bottomrule
\end{tabular}
\caption{Ablation on noise injection parameter $\lambda_\mathrm{noise}$.}
\label{table:prompt_perturbation}
\end{table}
Table \ref{table:prompt_perturbation} presents an ablation study on the noise injection parameter for the SAE loss. The results indicate that a parameter of 0.1 achieves the lowest structure distance, signifying the best preservation of the source content's geometry. Conversely, as the injection parameter increases, the CLIP score also rises. This suggests a trade-off: a larger parameter pushes the inverted latent to be more editable and better aligned with the target prompt, but at the cost of structural fidelity. Given that our primary objective is to perform edits while faithfully preserving the original structure, we set this parameter to 0.1 for our main experiments.

We note that the increase is primarily observed in CLIP 
Whole, while CLIP Edit remains largely stable across all 
values of $\lambda_\mathrm{noise}$ (ranging from 20.40 to 
20.42). This is because CLIP Edit measures similarity only 
within the edited region, which is already well-aligned with 
the target prompt at $\lambda_\mathrm{noise}{=}0.1$; 
further noise injection mainly affects the non-edited 
regions, which are captured by CLIP Whole.

\section{Encoder Training and Analysis of Inverted Latents}\label{sec:C}
\subsection{Training Details}\label{sec:C1}
We trained the inversion encoder for 72 hours using a batch size of 6. The total number of trained samples is 426,000. The weight for the SAE loss, $\lambda_{\mathrm{sae}}$, was set to 1. For optimization, we used the AdamW \cite{loshchilovdecoupled} optimizer with a learning rate of $1 \times 10^{-5}$, betas of (0.9, 0.999), an epsilon of $1 \times 10^{-8}$, and a weight decay of $1 \times 10^{-2}$. The loss curves during training are visualized in Figures~\ref{fig:mseloss} and~\ref{fig:saeloss}.

\subsection{Visualization and Analysis of Inverted Latents}\label{sec:C2}

We analyze the statistical and spectral properties of 
latents produced by our SAE-trained inversion encoder, 
comparing against pure Gaussian noise 
$z \sim \mathcal{N}(0, I)$.

\subsubsection{Element-wise Gaussianity.}
Table~\ref{tab:latent_stats} (top) reports normality 
diagnostics over $1000$ inverted latents. The Q--Q $R^2$ of 
0.999, KS statistic of 0.069, and near-zero skewness and 
kurtosis confirm that the marginal distribution is nearly 
indistinguishable from $\mathcal{N}(0,1)$, verifying 
compatibility with the diffusion prior.

\subsubsection{Spectral Analysis.}\label{sec:spectral}
\begin{figure}[t]
    \centering
    \includegraphics[width=0.5\linewidth]{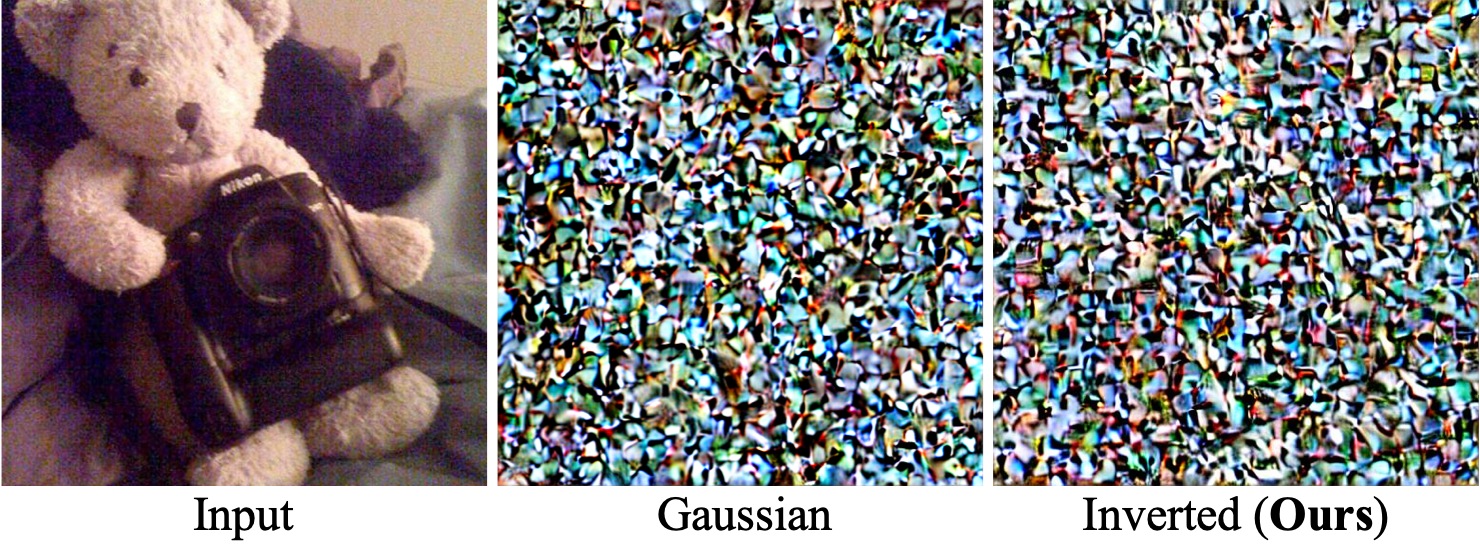}
    \caption{Visualization of inverted latents from our 
    encoder. Despite being element-wise Gaussian 
    (Table~\ref{tab:latent_stats}), the latents exhibit 
    visible low-frequency structure inherited from the 
    source image.}
    \label{fig:inverted_latent}
\end{figure}

Element-wise Gaussianity does not imply spatial 
independence. We compute the 2D power spectrum of each 
latent, average over channels, and extract radial 
profiles. We define the \textit{Low/High Ratio} as the 
mean power for $r < 0.15R$ divided by that for 
$r > 0.50R$. For spatially independent noise, this ratio 
is approximately 1. As shown in Table~\ref{tab:latent_stats}, 
our latents exhibit a ratio of 7.145---low-frequency 
energy is approximately \textbf{7$\times$ larger} than 
high-frequency energy, indicating that coarse spatial 
structure is implicitly encoded. This is also visually 
apparent in Figure~\ref{fig:inverted_latent}, where the 
inverted latent retains discernible low-frequency patterns 
from the source image despite appearing noise-like overall.

\subsubsection{Interpretation.}
Our inverted latents are element-wise Gaussian yet 
spatially structured: they appear as valid noise to the 
denoiser while encoding source-specific layout information 
in their spatial correlations. This dual property steers 
the denoising trajectory toward faithful reconstruction 
without violating the diffusion prior, explaining the 
superior inversion quality over DDIM reported in Table~1 
of the main paper.

\begin{table}[t]
\centering
\caption{Analysis of inverted latents. Top: element-wise 
Gaussianity. Bottom: Low/High Ratio measuring spatial 
structure in the power spectrum 
(Section~\ref{sec:spectral}).}
\label{tab:latent_stats}
\begin{tabular}{lcc}
    \toprule
    \textbf{Metric} & \textbf{Gaussian} & \textbf{Ours} \\
    \midrule
    Q--Q $R^2$           & 1.000 & 0.999 \\
    KS Statistic         & 0.000 & 0.069 \\
    Skewness             & 0.000 & $-$0.032 \\
    Kurtosis             & 0.000 & $-$0.083 \\
    \midrule
    Low/High Ratio       & -- & 7.145 \\
    \bottomrule
\end{tabular}
\end{table}

\section{Ablation of Anchor}\label{sec:D}
\begin{table}[h]
\fontsize{7.8pt}{9.2pt}\selectfont
\centering
\begin{tabularx}{1\textwidth}{l *{6}{>{\centering\arraybackslash}X}}
\toprule
\multirow[c]{2}{*}[-0.6ex]{Anchor method} & \multicolumn{6}{c}{90 Frames} \\
\cmidrule(lr){2-7}
& SC & BC & MS & AQ & IQ & BQS\\
\midrule
Random          &0.950&0.963&0.987&0.670&0.723&0.673 \\
First           &0.958&0.962&0.988&0.670&0.723&0.675 \\
Pixel-Centroid  &0.957&0.962&0.988&0.670&0.723&0.675 \\
Pixel-Medoid    &0.957&0.964&0.988&0.670&0.723&0.675 \\
CLIP Top-1      &0.957&0.963&0.988&0.670&0.723&0.675 \\
\rowcolor{gray!30} \textbf{DINO-Medoid}     &0.958&0.965&0.989&0.670&0.723&0.676 \\
Hybrid          &0.956&0.962&0.988&0.670&0.723&0.675\\
\bottomrule
\end{tabularx}
\caption{Ablation of anchor Methods (90 Frames).}
\label{tab:anchorabl}
\end{table}
\subsection{Anchor selection strategies ablation}\label{sec:D1}
To determine the optimal anchor selection strategy, we conducted an ablation study comparing our chosen method against six alternatives. The evaluated methods include: 1) \textit{Random}, which selects a random frame; 2) \textit{First}, which always uses the initial frame; 3) \textit{Pixel-Centroid}, which selects the frame closest to the pixel-wise mean of all frames; 4) \textit{Pixel-Medoid}, which selects the frame minimizing the sum of pixel-wise distances to all other frames; 5) \textit{DINO-Medoid}, which identifies the medoid in the DINO \cite{caron2021emerging} feature space; 6) \textit{CLIP Top-1}, which selects the frame with the highest CLIP \cite{radford2021clip} similarity to the source prompt; and 7) \textit{Hybrid}, which considers both DINO and CLIP scores. The ablation was performed on 90-frame videos to capture the long-term effects of the anchor choice. As shown in Table \ref{tab:anchorabl}, the DINO-Medoid strategy consistently yielded the best performance, achieving the highest scores across our metrics. Notably, the Hybrid strategy does not outperform DINO-Medoid despite combining more information. We conjecture that 
aggregating two heterogeneous scores introduces a weighting 
ambiguity: the selected frame becomes a compromise that does 
not maximize either criterion. Given that all non-random 
strategies already select reasonably representative frames 
and the absolute differences are small, a single well-chosen 
criterion (DINO feature similarity) proves sufficient and 
avoids the need for additional hyperparameter tuning to 
balance the two scores.

\subsection{Multi-Anchor Ablation}\label{sec:D2}
\begin{figure}[t]
    \centering
    \includegraphics[width=0.7\linewidth]{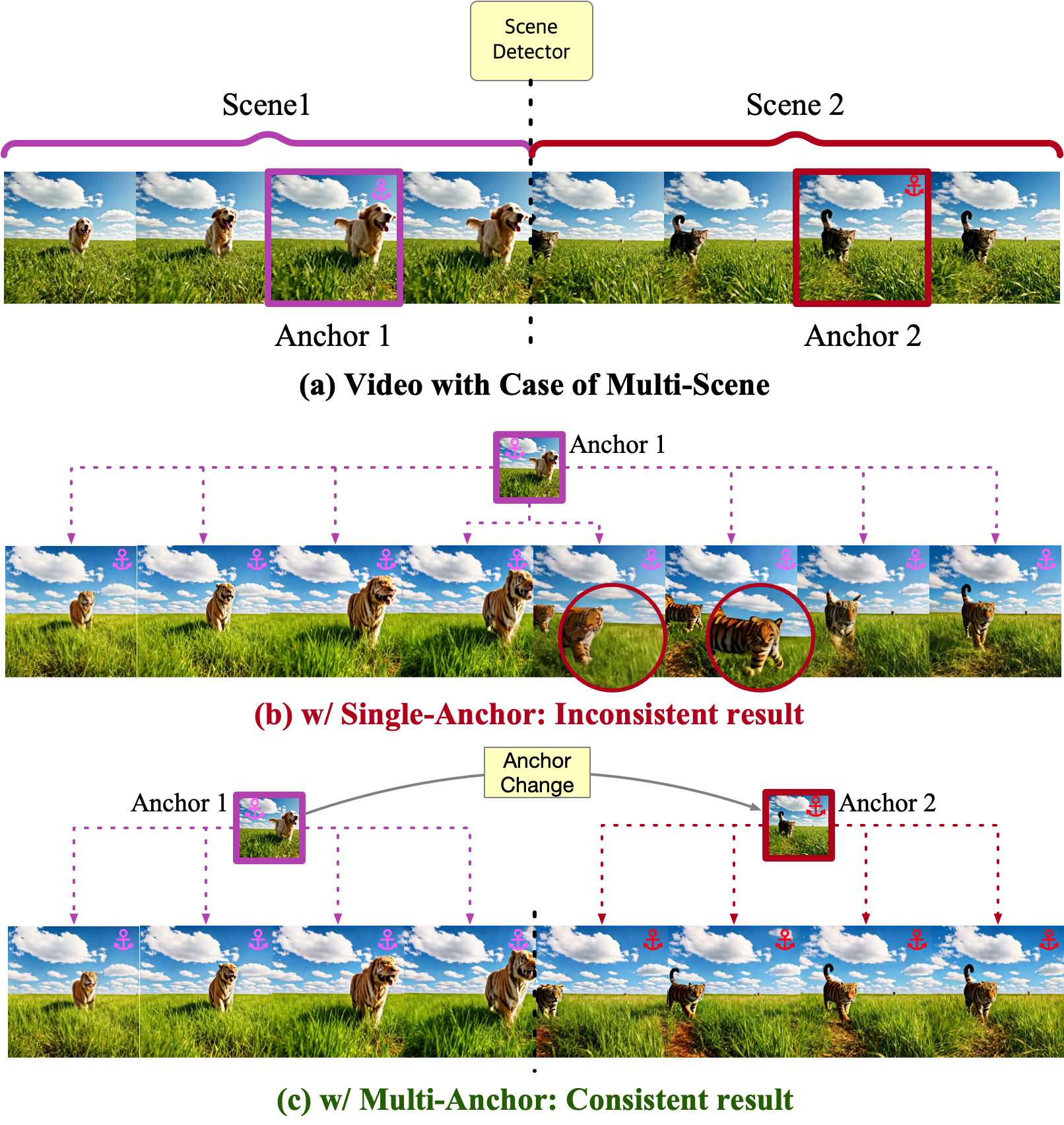}
    \caption{Overview of the multi-anchor strategy. (a) A multi-scene video example (dog to cat). (b) When using only a single anchor derived from the dog's scene, the editing quality of the cat's region degrades noticeably (see red circles). (c) By employing multiple anchors, our method preserves visual consistency for each object across different scenes.}
    \label{fig:multi-anchor}
\end{figure}

We further demonstrate that our single-anchor framework can be naturally extended to a multi-anchor strategy to handle more complex video scenarios. When a video contains only a single object or a single coherent scene, a single anchor is sufficient to maintain temporal consistency throughout the editing process. However, for videos composed of multiple distinct scenes—for example, a video that transitions from a cat walking to a dog walking (Figure~\ref{fig:multi-anchor}(a))—relying on a single anchor becomes problematic. In such cases, the anchor selected from one scene (e.g., the dog's scene) fails to provide meaningful correspondences for frames belonging to a different scene (e.g., the cat's scene), resulting in noticeable degradation in editing quality (Figure~\ref{fig:multi-anchor}(b)).
To address this limitation, we introduce a simple yet effective multi-anchor strategy. We first apply an off-the-shelf scene change detector (e.g., PySceneDetect\footnote{\url{https://github.com/Breakthrough/PySceneDetect}}) to segment the input video into individual scenes. We then independently select an anchor frame for each detected scene. During inference, anchors are dynamically swapped at scene boundaries so that each segment is always paired with its corresponding anchor. As shown in Figure~\ref{fig:multi-anchor}(c), this strategy enables our method to preserve per-scene consistency without compromising editing quality across scene transitions.
\section{Ablation on Sliding Window and Anchor}\label{sec:E}
In this section, we conduct an ablation study on the sliding window size and stride to find the optimal balance between editing quality and processing speed. As shown in Tables~\ref{tab:swaabl_wo} and~\ref{tab:swaabl_w}, a window size of 7 and a stride of 5 (with an anchor) provides a strong trade-off, which we adopt for our main experiments. 

We note that SC, BC, and MS consistently improve as the window 
size increases, confirming that larger windows provide stronger 
temporal context. In contrast, AQ exhibits a slight decrease 
(e.g., from 0.680 at $w{=}1$ to 0.670 at $w{=}7$ in 
Table~\ref{tab:swaabl_w}). This is because tiling more frames 
into a single grid reduces the effective per-frame resolution 
seen by the UNet, which marginally affects per-frame aesthetic 
quality. However, the magnitude of this decrease is negligible 
($\sim$0.01), and the substantial gains in temporal consistency 
metrics justify the trade-off.

Comparing the two tables, the anchor consistently improves 
consistency metrics (e.g., SC increases from 0.931 to 0.943 at 
$w{=}1$), but incurs an FPS reduction (e.g., 20.9 to 15.3 at 
$w{=}1$). This is expected: adding an anchor increases the 
number of grid slots from $w$ to $w{+}r$ 
(cf.\ Section~\ref{sec:H}), and the relative overhead is most 
pronounced at small window sizes where the anchor constitutes a 
larger fraction of the total grid.

\newcolumntype{C}{>{\centering\arraybackslash}X}
\begin{table}[t]
\fontsize{7.8pt}{9.2pt}\selectfont
\setlength{\tabcolsep}{3pt}
\renewcommand{\arraystretch}{1.12}
\centering

\resizebox{0.75\textwidth}{!}{%
\begin{tabularx}{0.75\textwidth}{@{} c c *{7}{C} @{}}
\toprule
\makecell{Window \\ size $w$} & Stride & SC & BC & MS & AQ & IQ & BQS & FPS \\
\midrule
1 & 1 & 0.931 & 0.948 & 0.980 & 0.680 & 0.723 & 0.668 & 20.925 \\
\cmidrule{1-9}
3 & 1 & 0.950 & 0.962 & 0.988 & 0.676 & 0.723 & 0.676 & 11.646 \\
\cmidrule{1-9}
\multirow[c]{2}{*}{5}
& 1 & 0.952 & 0.963 & 0.988 & 0.674 & 0.722 & 0.676 & 7.068 \\
& 3 & 0.952 & 0.962 & 0.987 & 0.673 & 0.722 & 0.675 & 17.409 \\
\cmidrule{1-9}
\multirow[c]{3}{*}{\textbf{7}}
& 1 & 0.954 & 0.964 & 0.988 & 0.670 & 0.722 & 0.674 & 4.769 \\
& 3 & 0.954 & 0.963 & 0.988 & 0.671 & 0.722 & 0.674 & 12.354 \\
& \cellcolor{gray!30}\textbf{5} & \cellcolor{gray!30}0.954 & \cellcolor{gray!30}0.963 & \cellcolor{gray!30}0.988 & \cellcolor{gray!30}0.672 & \cellcolor{gray!30}0.723 & \cellcolor{gray!30}0.675 & \cellcolor{gray!30}18.140 \\
\cmidrule{1-9}
\multirow[c]{4}{*}{9}
& 1 & 0.955 & 0.963 & 0.989 & 0.669 & 0.723 & 0.674 & 3.351 \\
& 3 & 0.954 & 0.962 & 0.988 & 0.670 & 0.723 & 0.674 & 9.203 \\
& 5 & 0.954 & 0.965 & 0.988 & 0.669 & 0.723 & 0.674 & 13.862 \\
& 7 & 0.955 & 0.964 & 0.988 & 0.670 & 0.723 & 0.674 & 17.599 \\
\bottomrule
\end{tabularx}%
}
\caption{Ablation on Unified-Frame Editing \textbf{without anchor} on 90-frame videos.}
\label{tab:swaabl_wo}

\vspace{0.5em}

\resizebox{0.75\textwidth}{!}{%
\begin{tabularx}{0.75\textwidth}{@{} c c *{7}{C} @{}}
\toprule
\makecell{Window \\ size $w$} & Stride & SC & BC & MS & AQ & IQ & BQS & FPS \\
\midrule
1 & 1 & 0.943 & 0.950 & 0.985 & 0.680 & 0.722 & 0.669 & 15.332 \\
\cmidrule{1-9}
3 & 1 & 0.958 & 0.966 & 0.988 & 0.670 & 0.723 & 0.679 & 8.698 \\
\cmidrule{1-9}
\multirow[c]{2}{*}{5}
& 1 & 0.957 & 0.965 & 0.988 & 0.670 & 0.723 & 0.678 & 5.644 \\
& 3 & 0.958 & 0.963 & 0.988 & 0.670 & 0.723 & 0.677 & 14.324 \\
\cmidrule{1-9}
\multirow[c]{3}{*}{\textbf{7}}
& 1 & 0.957 & 0.966 & 0.988 & 0.670 & 0.723 & 0.677 & 3.959 \\
& 3 & 0.957 & 0.964 & 0.988 & 0.670 & 0.723 & 0.676 & 10.551 \\
& \cellcolor{gray!30}\textbf{5} & \cellcolor{gray!30}0.958 & \cellcolor{gray!30}0.965 & \cellcolor{gray!30}0.989 & \cellcolor{gray!30}0.670 & \cellcolor{gray!30}0.723 & \cellcolor{gray!30}0.677 & \cellcolor{gray!30}15.793 \\
\cmidrule{1-9}
\multirow[c]{4}{*}{9}
& 1 & 0.957 & 0.963 & 0.989 & 0.670 & 0.723 & 0.675 & 2.870 \\
& 3 & 0.957 & 0.961 & 0.988 & 0.670 & 0.723 & 0.675 & 7.886 \\
& 5 & 0.957 & 0.964 & 0.988 & 0.670 & 0.723 & 0.676 & 12.109 \\
& 7 & 0.958 & 0.964 & 0.988 & 0.670 & 0.723 & 0.676 & 15.614 \\
\bottomrule
\end{tabularx}%
}
\caption{Ablation on Unified-Frame Editing \textbf{with anchor} on 90-frame videos.}
\label{tab:swaabl_w}

\end{table}
\section{Details of Unified-Frame Editing}\label{sec:F}
We prepare the sequence for sliding window processing. To ensure sufficient context for boundary frames, we apply reflect padding to the sequence of inverted latents $\mathcal{\hat V}_K^T = (\hat z^T_0, \dots, \hat z^T_{K-1})$. Let $p = (w-1)/2$ be the padding size (for an odd $w$). The padded sequence $\mathcal{\hat V}_{\text{pad}}^T$ is constructed as:
$$
\mathcal{\hat V}_{\text{pad}}^T = (\underbrace{\hat{z}^T_{p}, \dots, \hat{z}^T_1}_{\text{left reflection}}, \underbrace{\hat{z}^T_0, \dots, \hat{z}^T_{K-1}}_{\text{original}}, \underbrace{\hat{z}^T_{K-2}, \dots, \hat{z}^T_{K-1-p}}_{\text{right reflection}})
$$

The padded sequence is processed to generate a series of output segments. The total number of segments, $N_w$, is given by $N_w = \lceil K/s \rceil$. For each segment index $i \in \{0, 1, \dots, N_w-1\}$, we first extract a window of $w$ latents from the padded sequence. These are spatially concatenated to form the window map $W^T_i$:
$$
W^T_i = \mathrm{Concat}(\hat{z}^T_{i \cdot s}, \dots, \hat{z}^T_{i \cdot s + w - 1}, \text{axis}=2)
$$
The final input for the generator, $W'^T_i$, is then constructed by prepending the anchor latent $\hat{z}^T_A$ to this window map:
$$
W'^T_i = \mathrm{Concat}(\hat{z}^T_A, W^T_i, \text{axis}=2)
$$
The generator processes this input, producing a unified output map $W'^0_i = G_\theta(W'^T_{i}, c_{\mathrm{edit}})$. From this output, we first discard the initial portion corresponding to the anchor frame to get $W^0_i$:
$$
W^0_i = W'^0_i[:, :, W:] \quad\quad (\text{Dimensions: } \mathbb{R}^{C \times H \times (W \cdot w)})
$$
Next, we extract the central $s$ frames from $W^0_i$ to form the final segment $S_i$:
$$
S_i = W^0_i[:, :, W \cdot \text{offset} : W \cdot (\text{offset} + s)] \quad\quad (\text{where offset} = (w-s)/2)
$$
After generating all $N_w$ segments, they are concatenated. The resulting map is truncated to the original length of $K$ frames to form the final edited latent map, $Z^0_{\text{edit}}$:
$$
Z^0_{\text{edit}} = \mathrm{Truncate}(\mathrm{Concat}(S_0, S_1, \dots, S_{N_w-1}, \text{axis}=2), K).
$$
Finally, this unified map is partitioned back into a sequence of individual frame latents:
$$
\hat z^0_{k_\mathrm{edit}} = Z^0_{\text{edit}}[:, :, k \cdot W : (k+1) \cdot W] \quad \text{for } k=0, \dots, K-1.
$$
The final edited video, $\mathcal{V}^0_{\mathrm{edit}}$, is the sequence of these $K$ latents:
$$
\mathcal{\hat V}^0_{\mathrm{edit}} = (\hat z^0_{0_\mathrm{edit}}, \hat z^0_{1_\mathrm{edit}}, \dots, \hat z^0_{K-1_\mathrm{edit}}).
$$
\section{Additional Analysis}\label{sec:G}

\subsection{Detailed Ablation on Noise Parameter $\lambda_{\mathrm{noise}}$}\label{sec:G1}
Detailed quantitative results are reported in Table~\ref{tab:noise_std}.
As $\lambda_{noise}$ increases, both the CLIP~\cite{radford2021clip} 
distance (CLIP-D) and the structure distance~\cite{caron2021emerging} 
with respect to the original image increase, and their standard 
deviations also grow. Nevertheless, for moderate values of 
$\lambda_{noise}$ (e.g., $\lambda_{noise} = 0.1$), we can generate 
images that are semantically slightly different while remaining 
structurally similar to the original, with only a small variability 
across samples.

\subsection{Discussion on One-Step T2V Generator Quality}

As discussed in Section~5 (Discussion) of the main paper, we build OSVE 
on a one-step T2I backbone because one-step text-to-video 
(T2V) generators have not yet reached a level of quality 
sufficient for reliable video editing applications. In this 
section, we review the current state of accelerated T2V 
generation and discuss why one-step T2V remains an open 
and challenging problem.

\subsubsection{Current State of T2V Distillation.}
State-of-the-art T2V diffusion models such as 
CogVideoX~\cite{yang2024cogvideox}, 
HunyuanVideo~\cite{kong2024hunyuanvideo}, and 
Wan2.1~\cite{wan2025wanopenadvancedlargescale} typically 
require 30--50 denoising steps, resulting in generation 
times on the order of minutes even on high-end GPUs. 
While the image domain has seen remarkable progress in 
distillation---with one-step T2I generators such as 
DMD2~\cite{yin2024improved} and 
SDXL-Turbo~\cite{sauer2025adversarial} achieving quality 
on par with their multi-step teachers---extending these 
techniques to video has proven substantially more 
difficult. Recent efforts have explored trajectory-based 
approaches~\cite{salimans2022progressive,song2023consistency}, 
distribution matching 
methods~\cite{yin2024onestep,yin2024improved}, and their 
combinations~\cite{zheng2025rcm,berner2026tmd,yin2025causvid,zhang2025turbodiffusion}, 
but video distillation remains in a nascent stage where 
even few-step (2--8 step) methods exhibit measurable 
quality gaps relative to their full-step teachers in 
terms of temporal coherence, motion dynamics, text 
alignment, and generation 
diversity~\cite{zheng2025rcm,gpd2025,selfforcing2025pp}.

\label{sec:G2}
\begin{figure}
    \centering
    \includegraphics[width=0.9\linewidth]{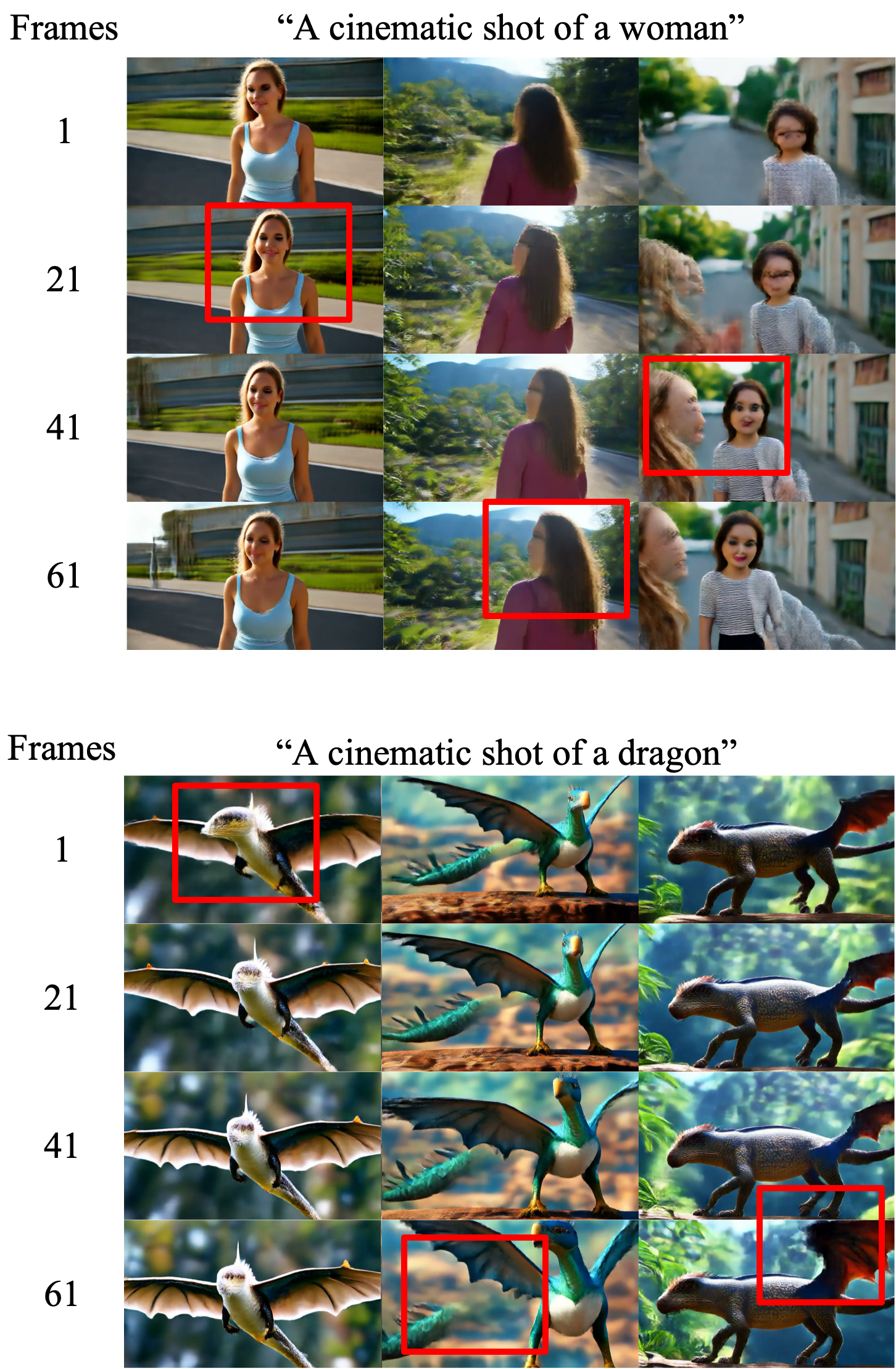}
    \caption{Qualitative results of one-step generation using Wan-rCM 1.3B. The outputs exhibit noticeable quality degradation, including spatial blurriness, temporal flickering, and discontinuous motion across frames, illustrating the fundamental limitations of current one-step T2V distillation discussed in this section.}
    \label{fig:WanrCM}
\end{figure}

\subsubsection{One-Step T2V: Fundamental Quality Limitations}
When pushed to the extreme of a single denoising step, 
the quality degradation becomes severe. The few works that 
have attempted one-step T2V generation consistently report 
fundamental limitations.

APT~\cite{lin2025apt} proposes adversarial post-training 
for one-step video generation, producing 2-second 720p 
videos in real time. However, the authors themselves 
acknowledge significant quality degradation, and the 
approach has been demonstrated only on short clips with 
limited resolution.
POSE~\cite{cheng2025pose} introduces phased adversarial 
equilibrium for one-step video distillation and identifies 
three fundamental limitations of existing approaches: 
(i) an efficiency bottleneck in scaling to large models 
($>$10B parameters), (ii) temporal consistency degradation 
including flickering artifacts and broken long-range 
dependencies, and (iii) limited task generalization to 
conditional downstream tasks.
OSV~\cite{mao2025osv} combines consistency distillation 
with GAN training for one-step image-to-video generation, 
but requires multi-step refinement to match the teacher's 
quality.
Even rCM~\cite{zheng2025rcm}, one of the strongest 
distillation frameworks to date, reports a VBench score 
gap of $\sim$2 points between its 1-step and 4-step 
variants on the same Wan2.1 backbone, and explicitly 
notes that the distilled model does not surpass the 
teacher in diversity and physical consistency.
We further corroborate this observation by running 
Wan-rCM 1.3B in the one-step setting; as shown in 
Figure~\ref{fig:WanrCM}, the generated frames suffer 
from severe blurriness and temporal discontinuities, 
confirming that one-step T2V generation remains far 
from practical quality.

These limitations stem from challenges that are unique 
to or amplified in the video domain:
\begin{itemize}
    \item \textbf{Temporal consistency collapse.} One-step 
    generators must produce temporally coherent motion in 
    a single forward pass, which often leads to flickering, 
    jittering, or static 
    outputs~\cite{cheng2025pose,yin2025causvid}.
    
    \item \textbf{Spatiotemporal mode averaging.} The mode 
    averaging problem---where the model averages over 
    conflicting denoising directions---is dramatically 
    amplified in the high-dimensional spatiotemporal 
    setting of video, with no opportunity for iterative 
    correction in one-step generation.
    
    \item \textbf{Training instability.} 
    POSE~\cite{cheng2025pose} observes that adversarial 
    distillation from Gaussian noise faces severe training 
    instability in the video domain due to the high 
    dimensionality of video latents, often resulting in 
    mode collapse or training divergence.
\end{itemize}

\subsubsection{Implications for Our Approach.}
The contrast between T2I and T2V distillation is stark: 
one-step T2I generators reliably match their multi-step 
teachers, whereas T2V distillation---at any step 
count---still exhibits notable quality gaps, with 
one-step T2V suffering from particularly severe 
degradation. Our OSVE therefore adopts a one-step T2I 
backbone combined with explicit temporal consistency 
mechanisms, which provides precise per-frame structural 
control while maintaining temporal coherence---a 
combination that current T2V distillation methods cannot 
reliably offer regardless of step count. As T2V 
distillation technology matures, future work may revisit 
this design choice.

\section{GPU Memory Analysis}\label{sec:H}
A key advantage of the sliding window formulation above is that GPU
memory consumption is determined solely by the window size $w$ and is
\emph{independent} of the total sequence length $K$, since only a
single window resides on the GPU at any time.

Concretely, let $r$ and $c$ denote the number of rows and columns
used to tile the $w$ latent codes in a window onto a 2D spatial grid,
so that $w = r \cdot c$. For each window, our implementation
constructs a single tiled latent tensor of shape
$C \times (r\,H_\ell) \times (c'\,W_\ell)$, where $H_\ell$ and
$W_\ell$ are the spatial dimensions of a single latent code, and feeds
it to the denoising UNet and the VAE decoder.

When no anchor is used, we have $c' = c$ and the number of latent
slots in the grid equals the window size:
\begin{equation}
  N_{\mathrm{eff}}(w) = r \cdot c = w.
\end{equation}
When an anchor latent $\hat{z}^T_A$ is prepended (cf.\ the
construction of $W'^T_i$), the grid must accommodate $w + 1$ latent
codes in $r$ rows, requiring
$c' = \lceil (w+1)/r \rceil$ columns. The effective number of grid
slots therefore becomes
\begin{equation}\label{eq:neff}
  N_{\mathrm{eff}}(w) = r \left\lceil \frac{w+1}{r} \right\rceil.
\end{equation}
Since $w = r \cdot c$ by construction, this simplifies to
$c' = c + 1$ and thus $N_{\mathrm{eff}}(w) = r(c+1) = w + r$.

The peak GPU memory can then be approximated as
\begin{equation}\label{eq:vram}
  M_{\mathrm{VRAM}}(w)
  \;\approx\;
  M_0 + \alpha \, N_{\mathrm{eff}}(w),
\end{equation}
where $M_0$ is a constant overhead (model weights, optimizer states,
etc.)\ and $\alpha$ is a per-slot cost that depends on the latent
resolution and the UNet architecture. Substituting the anchor case
gives
$M_{\mathrm{VRAM}}(w) = M_0 + \alpha\,(w + r)$, confirming that the
memory requirement grows \emph{linearly} in the window size,
$M_{\mathrm{VRAM}} = O(w)$.

Crucially, the total number of frames $K$ affects only the number of
windows $N_w = \lceil K/s \rceil$ and hence the \emph{processing
time}, not the peak memory. This allows our method to scale to
arbitrarily long videos without increasing GPU memory, as validated in
the ablation study (Tables~\ref{tab:swaabl_wo} and
\ref{tab:swaabl_w}).

\section{Details of User Study}\label{sec:I}

We recruited 26 participants anonymously through social media
platforms. No restrictions were placed on professional background,
and participants received no information about which method
corresponded to which result, ensuring an unbiased evaluation.

For each trial, a participant was shown the source video and
the editing results produced by six methods side by side.
The presentation order of the six methods was randomized
independently for every trial and every participant to eliminate
positional bias. Each participant evaluated three randomly sampled
videos from our datasets, yielding
$26 \times 3 \times 6 = 468$ individual ratings per criterion.

Participants were asked to rate each result on two criteria using
a 5-point Likert scale (1\,=\,Severe issues, 5\,=\,No issues):
\begin{itemize}
    \item \textbf{Temporal Consistency (TC):} whether the edited
    video maintains smooth and coherent motion without flickering
    or abrupt changes across frames.
    \item \textbf{Visual Quality (VQ):} whether the edited frames
    are visually pleasing and free from noticeable artifacts.
\end{itemize}

An example of the survey interface presented to participants is
shown in Figure~\ref{fig:user_study}.

\begin{figure}[t]
    \centering
    \includegraphics[width=\linewidth]{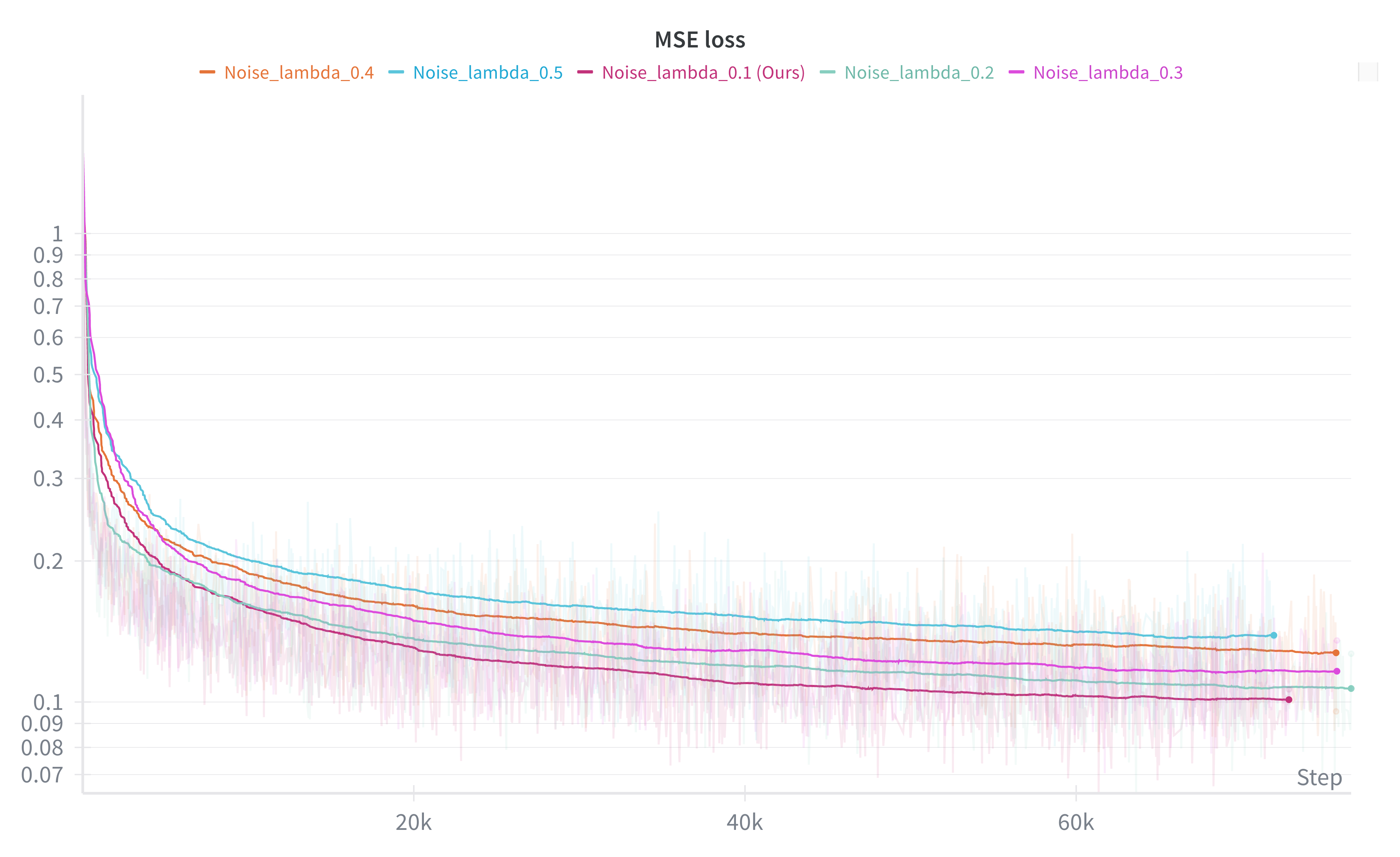}
    \caption{Graph of MSE loss while training.}
    \label{fig:mseloss}
\end{figure}
\begin{figure}[h]
    \centering
    \includegraphics[width=\linewidth]{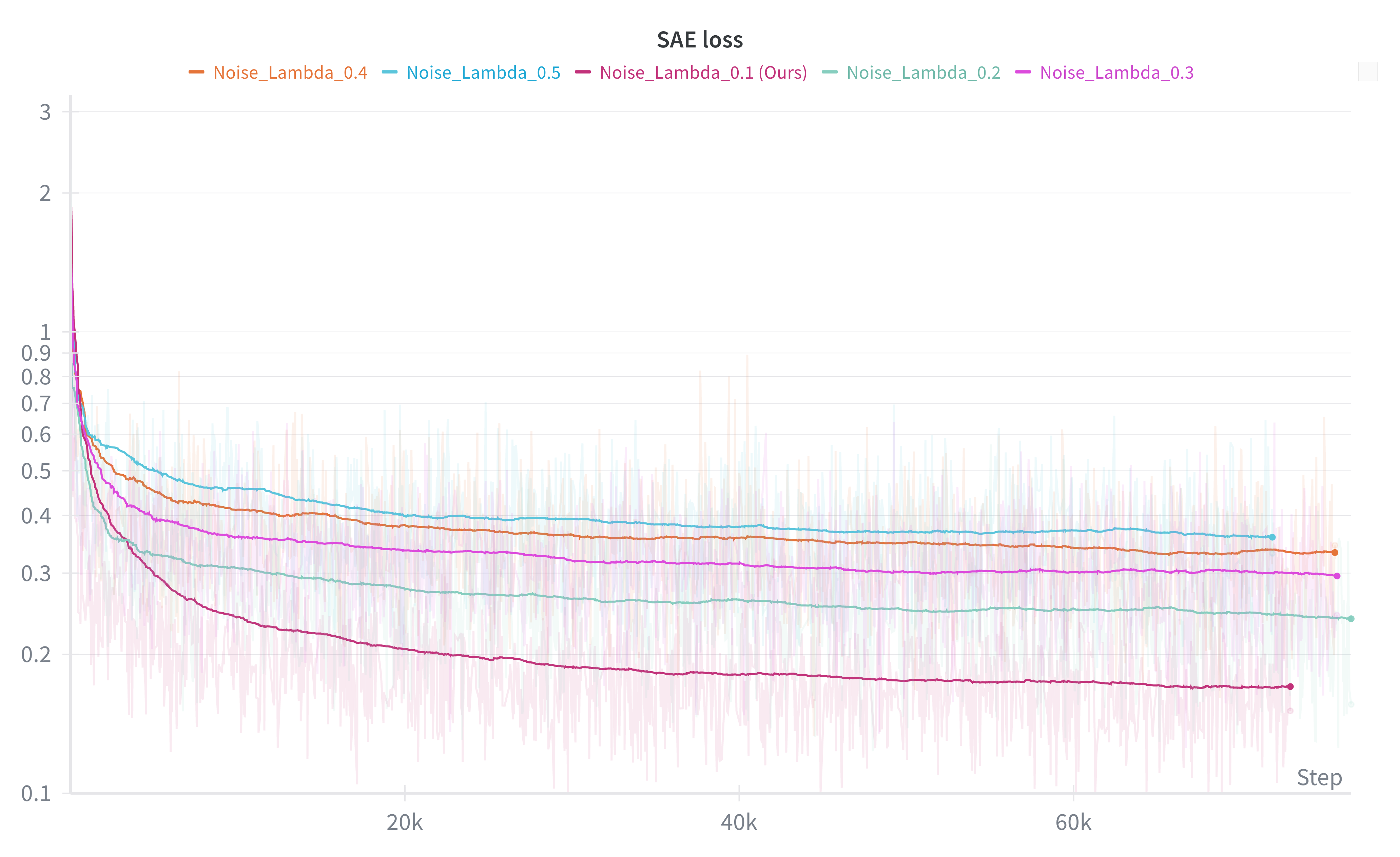}
    \caption{Graph of SAE loss while training.}
    \label{fig:saeloss}
\end{figure}
\clearpage

\begin{figure}
    \centering
    \includegraphics[width=0.5\linewidth]{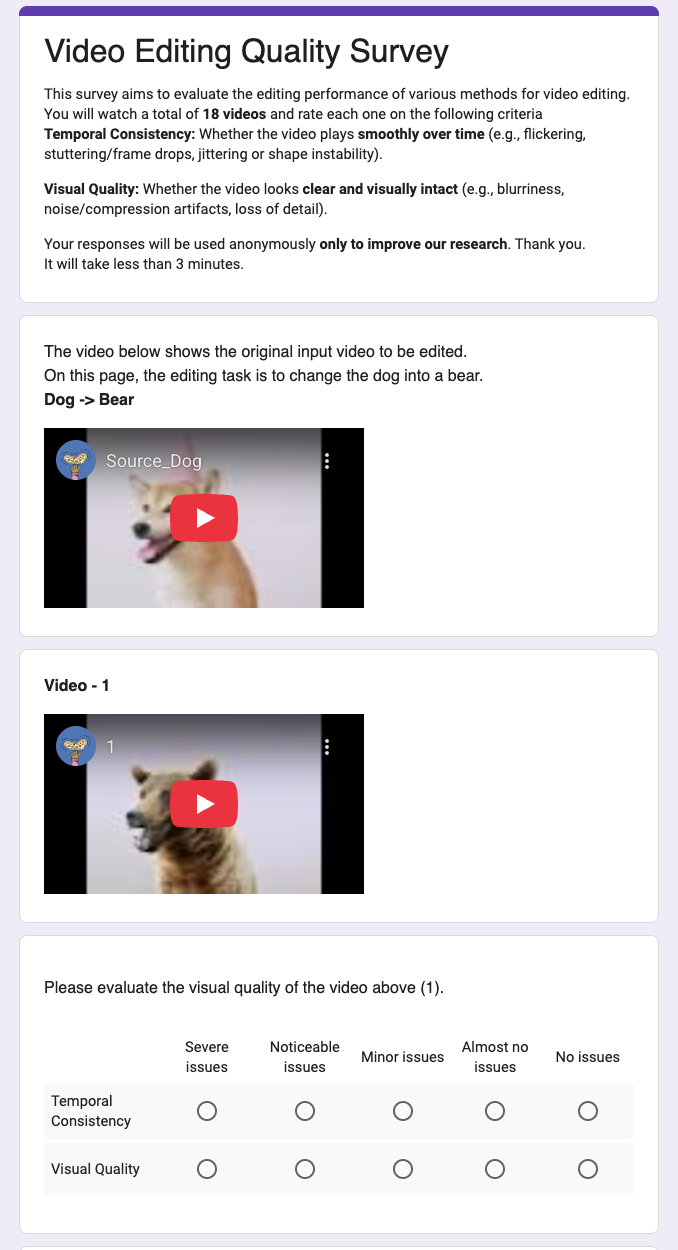}
    \caption{Screenshot of the survey interface presented to participants.}
    \label{fig:user_study}
\end{figure}

\clearpage

\begin{table*}[t]
\centering
\resizebox{0.4\textwidth}{!}{
\begin{tabular}{ccccc}
\toprule
$\lambda_{noise}$ &
\multicolumn{2}{c}{\shortstack{CLIP\\Distance (CLIP-D)}} &
\multicolumn{2}{c}{\shortstack{Structure\\Distance}} \\
& mean & std & mean & std \\
\midrule
0.000 & 1.00 & 0.000 & 0.000 & 0.000 \\
0.0100 & 0.994 & 0.00778 & 0.00230 & 0.00324 \\
0.0200 & 0.987 & 0.0139 & 0.00660 & 0.00779 \\
0.0300 & 0.981 & 0.0197 & 0.0106 & 0.0138 \\
0.0400 & 0.970 & 0.0321 & 0.0163 & 0.0206 \\
0.0500 & 0.968 & 0.0337 & 0.0227 & 0.0337 \\
0.0600 & 0.956 & 0.0392 & 0.0305 & 0.0393 \\
0.0700 & 0.958 & 0.0353 & 0.0301 & 0.0384 \\
0.0800 & 0.949 & 0.0439 & 0.0359 & 0.0358 \\
0.0900 & 0.944 & 0.0413 & 0.0374 & 0.0332 \\
0.100 & 0.944 & 0.0409 & 0.0415 & 0.0329 \\
0.110 & 0.936 & 0.0450 & 0.0414 & 0.0312 \\
0.120 & 0.931 & 0.0479 & 0.0501 & 0.0378 \\
0.130 & 0.925 & 0.0513 & 0.0533 & 0.0406 \\
0.140 & 0.921 & 0.0560 & 0.0554 & 0.0426 \\
0.150 & 0.915 & 0.0533 & 0.0601 & 0.0496 \\
0.160 & 0.911 & 0.0594 & 0.0711 & 0.0677 \\
0.170 & 0.909 & 0.0574 & 0.0714 & 0.0566 \\
0.180 & 0.903 & 0.0576 & 0.0713 & 0.0647 \\
0.190 & 0.898 & 0.0595 & 0.0856 & 0.0775 \\
0.200 & 0.897 & 0.0613 & 0.0803 & 0.0704 \\
0.210 & 0.886 & 0.0684 & 0.0839 & 0.0664 \\
0.220 & 0.891 & 0.0648 & 0.0885 & 0.0682 \\
0.230 & 0.882 & 0.0651 & 0.0831 & 0.0692 \\
0.240 & 0.866 & 0.0651 & 0.102 & 0.0762 \\
0.250 & 0.870 & 0.0701 & 0.0924 & 0.0783 \\
0.260 & 0.866 & 0.0704 & 0.102 & 0.0878 \\
0.270 & 0.862 & 0.0607 & 0.101 & 0.0770 \\
0.280 & 0.856 & 0.0721 & 0.107 & 0.0867 \\
0.290 & 0.852 & 0.0695 & 0.106 & 0.0764 \\
0.300 & 0.854 & 0.0690 & 0.111 & 0.0829 \\
0.310 & 0.840 & 0.0738 & 0.115 & 0.0843 \\
0.320 & 0.839 & 0.0710 & 0.112 & 0.0762 \\
0.330 & 0.829 & 0.0707 & 0.114 & 0.0855 \\
0.340 & 0.827 & 0.0729 & 0.117 & 0.0918 \\
0.350 & 0.818 & 0.0838 & 0.124 & 0.0911 \\
0.360 & 0.822 & 0.0786 & 0.111 & 0.0752 \\
0.370 & 0.817 & 0.0817 & 0.125 & 0.102 \\
0.380 & 0.809 & 0.0709 & 0.122 & 0.0868 \\
0.390 & 0.808 & 0.0824 & 0.128 & 0.0979 \\
0.400 & 0.778 & 0.0913 & 0.129 & 0.0839 \\
0.410 & 0.784 & 0.0818 & 0.140 & 0.114 \\
0.420 & 0.778 & 0.0889 & 0.141 & 0.109 \\
0.430 & 0.772 & 0.0815 & 0.135 & 0.0894 \\
0.440 & 0.772 & 0.0716 & 0.141 & 0.108 \\
0.450 & 0.763 & 0.0818 & 0.145 & 0.103 \\
0.460 & 0.752 & 0.0902 & 0.137 & 0.102 \\
0.470 & 0.767 & 0.0821 & 0.145 & 0.123 \\
0.480 & 0.744 & 0.0846 & 0.151 & 0.123 \\
0.490 & 0.726 & 0.0921 & 0.151 & 0.117 \\
0.500 & 0.734 & 0.0705 & 0.147 & 0.123 \\
\bottomrule
\end{tabular}
}
\caption{
Ablation on the noise parameter $\lambda_{noise}$. We report the mean and standard deviation of CLIP distance (CLIP-D, lower is better) and structure distance.
}
\label{tab:noise_std}
\end{table*}

\clearpage
\bibliographystyle{splncs04}
\bibliography{main}
\end{document}